\documentclass[10pt]{article}

\usepackage[utf8]{inputenc}
\usepackage[T1]{fontenc}

\makeatletter
\newif\ifTMLR
\IfFileExists{tmlr.sty}{\TMLRtrue}{\TMLRfalse}
\ifTMLR
  \usepackage{tmlr} 
\else
  \usepackage[margin=1in]{geometry}
  \usepackage[numbers]{natbib}
  \newcommand{\name}[1]{#1}
  \newcommand{\addr}[1]{#1}
  
\fi
\makeatother

\usepackage{microtype}
\usepackage{graphicx}
\usepackage{booktabs}
\usepackage{multirow}
\usepackage{wrapfig}
\usepackage{amsmath,amssymb,amsthm,mathtools}
\usepackage{bm}
\usepackage{xcolor}
\usepackage{enumitem}
\usepackage{hyperref}
\usepackage{algorithm}
\usepackage{algpseudocode}
\usepackage{framed}
\usepackage{thmtools}
\usepackage{etoolbox} 
\usepackage{hyphenat} 

\usepackage{tikz}
\usetikzlibrary{arrows.meta,positioning,calc,shapes.geometric}

\newcommand{\cM}{\mathcal{M}}  
\newcommand{\cG}{\mathcal{G}}  
\newcommand{\cX}{\mathcal{X}} 

\title{PAC Guarantees for Reinforcement Learning:\\
Sample Complexity, Coverage, and Structure}

\author{\name Joshua Steier\thanks{\texttt{joshsteier@gmail.com}} \\ \addr Independent Researcher}

\newtheorem{definition}{Definition}

\newtheorem{theorem}{Theorem}

\theoremstyle{remark}

\newcommand{\E}{\mathbb{E}}

\newcommand{\R}{\mathbb{R}}

\newcommand{\cO}{\mathcal{O}}
\newcommand{\tO}{\tilde{\mathcal{O}}}
\newcommand{\tTheta}{\tilde{\Theta}}
\newcommand{\tOmega}{\tilde{\Omega}}
\newcommand{\cS}{\mathcal{S}}
\newcommand{\cA}{\mathcal{A}}
\newcommand{\cF}{\mathcal{F}}
\newcommand{\cH}{\mathcal{H}}
\newcommand{\cD}{\mathcal{D}}

\begin{document}
\maketitle

\begin{abstract}
When data is scarce or mistakes are costly, average-case metrics fall short.
What a practitioner needs is a guarantee: with probability at least $1-\delta$, the learned policy is $\varepsilon$-close to optimal after $N$ episodes.
This is the PAC promise, and between 2018 and 2025 the RL theory community made striking progress on when such promises can be kept.
We survey that progress.
Our organizing tool is the Coverage-Structure-Objective (CSO) framework, proposed here, which decomposes nearly every PAC sample complexity result into three factors: coverage (how data were obtained), structure (intrinsic MDP or function-class complexity), and objective (what the learner must deliver).
CSO is not a theorem but an interpretive template that identifies bottlenecks and makes cross-setting comparison immediate.
The technical core covers tight tabular baselines and the uniform-PAC bridge to regret; structural complexity measures (Bellman rank, witness rank, Bellman-Eluder dimension) governing learnability with function approximation; results for linear, kernel/NTK, and low-rank models; reward-free exploration as upfront coverage investment; and pessimistic offline RL where inherited coverage is the binding constraint.
We provide practitioner tools: rate lookup tables indexed by CSO coordinates, Bellman residual diagnostics, coverage estimation with deployment gates, and per-episode policy certificates.
A final section catalogs open problems, separating near-term targets from frontier questions where coverage, structure, and computation tangle in ways current theory cannot resolve.
\end{abstract}

\paragraph{Keywords}
PAC RL; uniform--PAC; sample complexity; function approximation; offline RL; low--rank MDPs; Bellman--Eluder dimension.


\section{Introduction}
\label{sec:introduction}

\begin{figure}[t]
\centering
\begin{tikzpicture}[scale=0.9, every node/.style={font=\small}]
  \node[draw, rectangle] (intro) at (0,0) {§1 Intro};
  \node[draw, rectangle] (tax) at (3,0) {§2 CSO};
  \node[draw, rectangle] (prelim) at (6,0) {§3 Prelim};
  \node[draw, rectangle] (tab) at (2,-1.5) {§4 Tabular};
  \node[draw, rectangle] (struct) at (5,-1.5) {§5 Structure};
  \node[draw, rectangle] (func) at (8,-1.5) {§6 Function};
  \node[draw, rectangle] (rich) at (5,-3) {§7 Rich Obs};
  \node[draw, rectangle] (rfe) at (8,-3) {§8 RFE};
  \node[draw, rectangle] (offline) at (2,-3) {§9 Offline};
  \node[draw, rectangle] (prac) at (5,-4.5) {§14 Practice};
  \draw[->] (intro) -- (tax);
  \draw[->] (tax) -- (prelim);
  \draw[->] (prelim) -- (tab);
  \draw[->] (prelim) -- (struct);
  \draw[->] (prelim) -- (func);
  \draw[->] (tab) -- (offline);
  \draw[->] (struct) -- (func);
  \draw[->] (struct) -- (rich);
  \draw[->] (func) -- (rfe);
  \draw[->] (func) -- (rich);
  \draw[->] (offline) -- (prac);
  \draw[->] (rich) -- (prac);
  \draw[->] (rfe) -- (prac);
\end{tikzpicture}
\caption{\textbf{Survey roadmap.} The CSO framework (§2) organizes all results. Preliminaries (§3) introduce formal tools. Three pillars (tabular §4, structural measures §5, function approximation §6) feed into applications (§7--§9). The practical toolkit (§14) synthesizes. \textbf{Reading paths:} Practitioners start at §2 and §14, then read domain sections; theorists start at §3, then §4--§6.}
\label{fig:roadmap}
\end{figure}
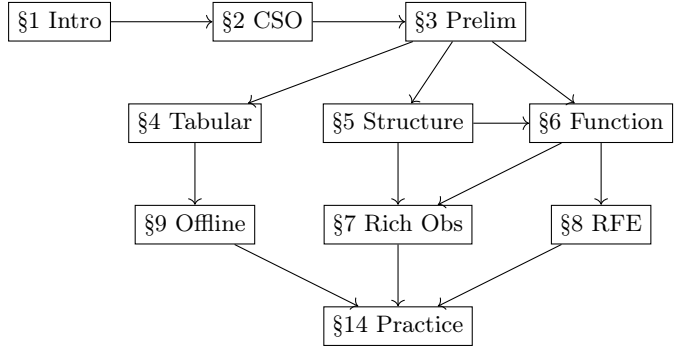

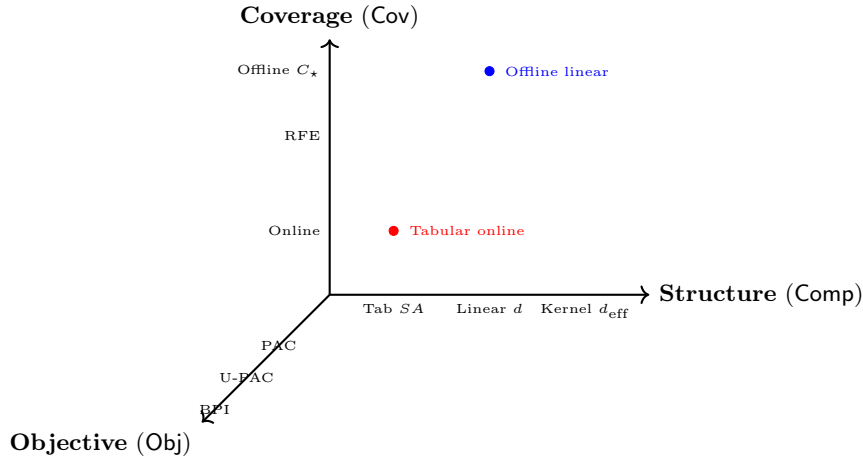
\begin{figure}[t]
\centering
\begin{tikzpicture}[scale=0.85, every node/.style={font=\small}]
  \draw[->,thick] (0,0) -- (5,0) node[right] {\textbf{Structure} ($\mathsf{Comp}$)};
  \draw[->,thick] (0,0) -- (0,4) node[above] {\textbf{Coverage} ($\mathsf{Cov}$)};
  \draw[->,thick] (0,0) -- (-2,-2) node[below left] {\textbf{Objective} ($\mathsf{Obj}$)};
  \node[below,font=\tiny] at (1,0) {Tab $SA$};
  \node[below,font=\tiny] at (2.5,0) {Linear $d$};
  \node[below,font=\tiny] at (4,0) {Kernel $d_{\mathrm{eff}}$};
  \node[left,font=\tiny] at (0,1) {Online};
  \node[left,font=\tiny] at (0,2.5) {RFE};
  \node[left,font=\tiny] at (0,3.5) {Offline $C_\star$};
  \node[font=\tiny] at (-0.8,-0.8) {PAC};
  \node[font=\tiny] at (-1.3,-1.3) {U-PAC};
  \node[font=\tiny] at (-1.8,-1.8) {BPI};
  \filldraw[blue] (2.5,3.5) circle (2pt);
  \node[right,font=\tiny,blue] at (2.6,3.5) {Offline linear};
  \filldraw[red] (1,1) circle (2pt);
  \node[right,font=\tiny,red] at (1.1,1) {Tabular online};
\end{tikzpicture}
\caption{\textbf{The CSO space.} Every PAC guarantee occupies a point in Coverage $\times$ Structure $\times$ Objective space. Moving along an axis changes one factor in the sample complexity decomposition~\eqref{eq:cso}. Two example results are marked.}
\label{fig:cso}
\end{figure}

\paragraph{How to read this survey.}
Readers mainly interested in \emph{what to use when} can start with Table~\ref{tab:taxonomy},
the CSO template (\S\ref{sec:cso}), and the practical decision tree (\S\ref{sec:practice}), then jump to the relevant setting:
tabular baselines (\S\ref{sec:tabular}), structural complexity measures (\S\ref{sec:structure}), function approximation (\S\ref{sec:function}),
rich observations and low rank (\S\ref{sec:rich}), reward-free exploration (\S\ref{sec:rfe}), offline RL (\S\ref{sec:offline}),
partial observability (\S\ref{sec:pomdp}), or PAC-Bayes (\S\ref{sec:pacbayes}).
Readers seeking technical depth should read \S\ref{sec:preliminaries} and the definition blocks at the start of each section.
We keep proofs informal (dependencies and scalings) and point to primary sources for exact constants.
Figures~\ref{fig:roadmap} and \ref{fig:cso}, together with Tables~\ref{tab:taxonomy} and \ref{tab:structure-compare}, provide a quick triage guide.

\subsection{Reinforcement learning and the need for fixed-confidence guarantees}

In reinforcement learning (RL), an agent interacts with an environment over discrete time steps.
At each step, the agent observes a state $s$ (for instance, a robot's sensor readings or a patient's vital signs), selects an action $a$ (a motor command or a treatment choice), receives a scalar reward $r$, and transitions to a new state $s'$.
The agent's goal is to learn a \emph{policy}, a rule $\pi$ mapping states to actions, that accumulates as much reward as possible over a planning horizon $H$.
The central tension is between \emph{exploration} (trying unfamiliar actions to learn their consequences) and \emph{exploitation} (repeating actions already known to work well).
Foundational treatments of this problem appear in Sutton and Barto~\citep{suttonbarto2018book}, Szepesvári~\citep{szepesvari2010algorithms}, and Kaelbling, Littman, and Moore~\citep{kaelbling1996survey}.

Most RL research measures performance through \emph{regret}: the cumulative loss from playing suboptimal actions while learning.
Regret is an average-case metric, well suited to settings where occasional mistakes are tolerable and the agent will interact with the environment for a long time.
But many real problems do not fit this mold.
A clinical trial cannot afford to assign a harmful treatment arm for thousands of episodes while the algorithm converges.
An autonomous vehicle controller must perform reliably from the outset, not merely on average over its lifetime.
An offline policy learned from hospital records should not be deployed unless we can certify, with quantified confidence, that it improves on the status quo.

These settings call for \emph{fixed-confidence} guarantees.
Formally, an algorithm is $(\varepsilon,\delta)$-PAC (Probably Approximately Correct) if, with probability at least $1-\delta$, it outputs a policy whose value is within $\varepsilon$ of optimal after a specified number of episodes $N(\varepsilon,\delta)$.
The PAC framework originated in computational learning theory with Valiant's work on supervised classification~\citep{valiant1984pac} and was adapted to sequential decision-making by Strehl, Li, and Littman~\citep{strehl2009pacjmlr}, who established the first systematic PAC analyses for finite Markov decision processes (MDPs).
This survey asks three questions: what fixed-confidence guarantees are known for RL, which assumptions make them possible, and how do the guarantees scale with problem parameters?

\subsection{Scope and time window}

We focus on the period 2018 to 2025 because three developments converged to reshape the PAC RL landscape during this window.

First, the \emph{uniform-PAC} framework of Dann, Lattimore, and Brunskill~\citep{uniformpac2017} showed that PAC and regret are not competing objectives but two views of the same statistical challenge.
A uniform-PAC algorithm bounds the number of $\varepsilon$-suboptimal episodes for all accuracy levels $\varepsilon$ simultaneously, and this automatically implies high-probability regret bounds (Theorem~\ref{thm:uniformpac-regret}).
The bridge crystallized in late 2017 and set the stage for everything that followed.

Second, \emph{structural complexity measures} matured between 2017 and 2021.
Jiang et al.~\citep{jiang2017olive} introduced Bellman rank, Sun et al.~\citep{sun2019witness} proposed witness rank, and Jin, Liu, and Miryoosefi~\citep{jin2021bellmaneluder} defined the Bellman-Eluder dimension.
Together, these measures delineate when function approximation admits polynomial sample complexity, replacing the brute-force dependence on the number of states with problem-dependent capacity parameters.

Third, \emph{coverage-centric} formulations reshaped how we think about data access.
Jin et al.~\citep{jin2020rewardfree} formalized reward-free exploration, where the agent builds a reusable dataset before any reward is specified.
Jin, Yang, and Wang~\citep{jin2021pevi} and Shi et al.~\citep{shi2022pql} developed pessimistic offline RL with explicit dependence on how well the training data covers the states that matter for the target policy.
Li et al.~\citep{li2024aos} settled the minimax sample complexity for model-based offline learning in tabular settings, and Zhang et al.~\citep{zhang2024settling} resolved the online tabular case.

Pre-2018 foundations are well covered by existing references: Strehl, Li, and Littman~\citep{strehl2009pacjmlr} for classical PAC-MDP analysis, Szepesvári~\citep{szepesvari2010algorithms} for algorithmic and analytical foundations, Kaelbling, Littman, and Moore~\citep{kaelbling1996survey} for an early broad survey, and Sutton and Barto~\citep{suttonbarto2018book} for updated algorithmics.
We build on these works rather than recapitulating them.

\subsection{Core concepts in brief}

The formal apparatus is developed in \S\ref{sec:preliminaries}; here we sketch the key ideas to orient the reader.

A finite-horizon MDP $M = (\cS, \cA, P, r, H, \rho)$ has state space $\cS$ with $|\cS| = S$, action space $\cA$ with $|\cA| = A$, stage-dependent transitions and rewards, horizon $H$, and initial distribution $\rho$.
The optimal policy $\pi^\star$ maximizes cumulative reward; the gap $V_1^\star(\rho) - V_1^{\hat\pi}(\rho)$ measures the suboptimality of a candidate $\hat\pi$.

An $(\varepsilon,\delta)$-PAC algorithm guarantees that this gap is at most $\varepsilon$ with probability $1 - \delta$, after $N(\varepsilon,\delta)$ episodes (Definition~\ref{def:pac-prelim}).
Uniform-PAC strengthens this to hold for all $\varepsilon$ at once (Definition~\ref{def:uniform-pac}), which implies high-probability regret (Theorem~\ref{thm:uniformpac-regret}).

Beyond the tabular regime, sample complexity depends on structural properties of the MDP or function class.
The relevant parameters include feature dimension $d$ (linear models), Bellman rank $B$, witness rank $W$, Bellman-Eluder dimension $d_{\mathrm{BE}}$, effective dimension $d_{\mathrm{eff}}(\lambda)$ (kernel models), and latent rank $r$ (low-rank MDPs).
Each of these replaces the tabular factor $SA$ in the sample complexity expression, and the hierarchy among them (Figure~\ref{fig:structure-hierarchy}, Table~\ref{tab:structure-compare}) is a recurring theme.

In offline RL, where the agent cannot collect new data, the \emph{concentrability coefficient} $C_\star$ (Definition~\ref{def:concentrability}) quantifies how well the training distribution covers the states visited by the target policy.
Large $C_\star$ inflates sample requirements regardless of structural simplicity; when $C_\star$ is infinite, consistent policy improvement is impossible without further assumptions.

\subsection{What this survey contributes}

Four outputs distinguish this survey from prior treatments.

The first is the \textbf{CSO framework} (\S\ref{sec:cso}), a reading lens proposed in this survey that decomposes every PAC bound into Coverage, Structure, and Objective factors.
The CSO decomposition is not a theorem; it is an organizational principle that makes cross-setting comparison systematic and identifies which factor is the bottleneck when a guarantee is vacuous.
We develop the framework in detail in \S\ref{sec:cso} and use it to navigate the rest of the paper.

The second is a \textbf{unified technical synthesis} across settings that are typically treated in isolation.
We state results for tabular, linear, kernel, low-rank, reward-free, and offline settings using common notation, with explicit parameter dependencies and pointers to primary sources for exact constants.
Table~\ref{tab:taxonomy} cross-references every major result by its CSO coordinates.

The third is a set of \textbf{practitioner tools}: diagnostic procedures for checking realizability and Bellman completeness (Algorithm~\ref{alg:bellman}), coverage estimation algorithms with finite-sample deployment gates (Algorithm~\ref{alg:coveragegate}), and policy certificates that provide per-episode accountability (Definition~\ref{def:cert}).
These tools operationalize the theory for applied researchers who need to decide whether a PAC guarantee is relevant to their problem and whether their data support deployment.

The fourth is a \textbf{structured inventory of open problems} (\S\ref{sec:open}), organized by time horizon (near-term vs.\ frontier) and annotated with impossibility results that clarify where progress requires new ideas rather than incremental refinement.

\subsection{Canonical results that anchor the survey}

To give the reader early orientation, we highlight four results that recur throughout the survey and illustrate the range of the CSO template.
Precise statements and proofs appear in the referenced sections; here we emphasize the dependencies and what they reveal.

The first anchor is the \textbf{tabular minimax rate}.
Zhang et al.~\citep{zhang2024settling} confirmed that learning an $\varepsilon$-optimal policy in a finite MDP with $S$ states, $A$ actions, and horizon $H$ requires $\tTheta(SAH^3/\varepsilon^2)$ episodes (Theorem~\ref{thm:tabular-minimax}).
This is the sharpest known PAC bound in any RL setting, and every structured result must recover it as a special case.
Through the CSO lens, coverage is trivial (the agent explores online), structure is $SA$, and the objective is standard PAC control.

The second is the \textbf{uniform-PAC to regret bridge}.
Dann, Lattimore, and Brunskill~\citep{uniformpac2017} showed that any algorithm satisfying uniform-PAC with budget $N(\varepsilon,\delta)$ achieves high-probability regret $\cO(\int_0^H \min\{K, N(\varepsilon,\delta)\}\, d\varepsilon)$ after $K$ episodes (Theorem~\ref{thm:uniformpac-regret}).
When $N$ has polynomial dependence on $(S,A,H,1/\varepsilon)$, this recovers near-minimax regret.
The bridge means that PAC analyses and regret analyses inform each other, and we invoke this conversion repeatedly.

The third anchor is \textbf{function approximation with linear features}.
Jin, Yang, and Wang~\citep{jin2020linear} showed that under linear realizability with dimension $d$, optimistic least-squares value iteration achieves PAC sample complexity $\tO(d^3 H^4/\varepsilon^2)$ (Theorem~\ref{thm:linmdp-lsvi}).
The horizon exponent rises from $H^3$ to $H^4$ because estimation errors are correlated across states sharing features.
He, Zhou, and Gu~\citep{he2021uniformpaclinear} extended this to the uniform-PAC setting under Bellman completeness.
The Bellman-Eluder dimension of Jin, Liu, and Miryoosefi~\citep{jin2021bellmaneluder} generalizes further, providing the broadest sufficient condition for learnability with rich function classes (Theorem~\ref{thm:be-structure}).

The fourth is \textbf{offline RL via pessimism}.
When learning from a fixed dataset with no further interaction, the binding constraint shifts from exploration to coverage.
Shi et al.~\citep{shi2022pql} showed that pessimistic Q-learning under linear realizability and concentrability $C_\star$ achieves control error $\varepsilon$ with $\tO(\mathrm{poly}(d, C_\star, H)/\varepsilon^2)$ samples (Theorem~\ref{thm:offline-lin}).
Li et al.~\citep{li2024aos} settled the model-based tabular case at the minimax level.
In the CSO template, coverage $\mathrm{poly}(C_\star)$ replaces the trivial online factor of $1$, and poor data support can make guarantees vacuous no matter how simple the structure.

\subsection{Connections across settings}

Several threads tie the survey together and are worth flagging before the technical development begins.

The uniform-PAC framework is not merely a theoretical convenience; it is a translation device.
Any uniform-PAC result in \S\ref{sec:tabular}--\S\ref{sec:function} automatically yields a regret bound, and conversely, regret analyses that are tight enough can be converted back to PAC guarantees.
This bidirectional link means that a reader interested in regret can still extract value from PAC-focused sections, and vice versa.

Coverage appears in two guises that are conceptually dual.
In reward-free exploration (\S\ref{sec:rfe}), the agent \emph{constructs} coverage as a reusable resource, paying an upfront cost (an extra factor of $S$ in the tabular case) to support arbitrary downstream rewards.
In offline RL (\S\ref{sec:offline}), coverage is \emph{inherited} from the behavior policy and quantified by $C_\star$; the agent cannot improve it and must instead protect against its limitations through pessimism.
Recognizing this duality clarifies when to invest in exploration and when to accept the data as given.

Structural complexity measures form a strict hierarchy (Figure~\ref{fig:structure-hierarchy}).
Tabular rates are the tightest, linear MDP rates relax to dimension $d$, low-rank and bilinear classes generalize further, and the Bellman-Eluder dimension provides the broadest umbrella.
Moving along this hierarchy trades tighter constants for wider applicability; a practitioner's job is to locate their problem at the most specific level that fits, extracting the tightest available guarantee.

Instance-dependent identification (\S\ref{sec:tabular}) offers a different kind of improvement: replacing the worst-case structural factor with gap-weighted sums that can be dramatically smaller when the optimal policy is well separated from alternatives.
Extending this to function approximation settings is one of the open problems we highlight in \S\ref{sec:open}.

\subsection{Practical implications for applied researchers}

Several actionable lessons emerge from the theory, and we state them here so that practitioners can keep them in mind while reading the technical sections.

Before applying any PAC guarantee, verify that the underlying assumptions hold.
Linear MDP bounds assume that rewards and transitions are linear in known features; invoking these bounds with arbitrary deep network features violates realizability and can produce misleading confidence.
The Bellman residual diagnostic in \S\ref{subsec:misspec} (Algorithm~\ref{alg:bellman}) provides a concrete test: fit value functions via ridge regression on random-policy data and check whether held-out Bellman residuals grow with the horizon.
If they do, the linear class is misspecified and richer features or kernel methods are needed.

In multi-task settings where the reward function is not fixed in advance, reward-free exploration (\S\ref{sec:rfe}) offers a principled alternative to ad hoc data collection.
The upfront exploration budget creates a dataset that supports $\varepsilon$-optimal planning for any reward, amortizing the cost across tasks.

In offline pipelines, coverage is the gatekeeper.
Estimate coverage proxies (density ratios, ridge leverage scores) using the procedures in \S\ref{subsec:coverage} and apply the deployment gate of Algorithm~\ref{alg:coveragegate} before committing to policy improvement.
When coverage is poor, prefer off-policy evaluation (interval estimates rather than point improvement) and abstain from deployment until additional data can be collected.

Policy certificates (Definition~\ref{def:cert}), introduced by Dann et al.~\citep{dann2019certificates}, provide per-episode accountability: a data-dependent bound on suboptimality that can be monitored in real time.
Deploy only when the certificate falls below the desired tolerance; if it remains large, the algorithm needs more data.

\subsection{Open problems at a glance}

We close the introduction with a brief preview of the open problems developed in \S\ref{sec:open}, organized around the CSO axes.

On the structure axis, the most pressing gap is uniform-PAC guarantees for kernel and over-parameterized function classes under conditions that can be verified from data rather than assumed a priori.
Current results require Bellman completeness, which is rarely checkable in practice; the target is algorithms whose sample complexity scales with the effective dimension $d_{\mathrm{eff}}(\lambda)$ and whose completeness assumptions are validated by residual tests.

On the coverage axis, offline RL under model misspecification remains poorly understood.
Existing guarantees assume exact realizability; when the function class is misspecified, error decomposes into approximation, estimation, and coverage components, but sharp characterizations of this three-way tradeoff with tractable algorithms are lacking.

At the intersection of structure and objective, instance-dependent identification with function approximation is almost entirely unexplored.
Gap-based best-policy identification for linear or low-rank models, matching the tabular lower bounds weighted by reachability, is a natural and apparently feasible next step.

Finally, data-driven structure selection (choosing among linear, kernel, and low-rank classes based on observed data, with oracle inequalities and uniform-PAC guarantees) would bring the theoretical landscape closer to practical deployment, where the correct structural assumption is rarely known in advance.

\subsection{How this survey differs from prior work}

Table~\ref{tab:compare-surveys} positions this survey relative to the two closest predecessors.
Szepesvári's monograph~\citep{szepesvari2010algorithms} provides a deep algorithmic treatment of RL through approximately 2010, covering value iteration, temporal-difference learning, and policy gradient methods with both PAC and regret analyses, primarily in the tabular setting.
Strehl, Li, and Littman~\citep{strehl2009pacjmlr} established the PAC-MDP framework for finite MDPs and derived the first systematic sample complexity bounds, setting the classical baselines that all subsequent work builds on.

Our survey starts where these works leave off.
We cover the 2018--2025 period, during which uniform-PAC, structural complexity measures, reward-free exploration, and pessimistic offline RL transformed the field.
The CSO framework, proposed here, provides a cross-cutting organizational principle that neither predecessor offers.
And the practitioner tools (coverage gates, residual diagnostics, certificate-based deployment rules) are new contributions aimed at bridging the gap between theory and applied use.

\begin{table}[t]
\centering
\small
\setlength{\tabcolsep}{5pt}
\begin{tabular}{p{0.19\textwidth} p{0.26\textwidth} p{0.26\textwidth} p{0.26\textwidth}}
\toprule
& \textbf{Szepesvári (2010)} & \textbf{Strehl et al.\ (2009)} & \textbf{This survey (2018--2025)} \\
\midrule
\textbf{Venue/Type} & Monograph (Morgan \& Claypool) & JMLR survey article & TMLR survey (2025) \\
\textbf{Scope} & Algorithms for RL; primarily tabular and basic function approximation & PAC-MDP framework; finite MDPs; sample complexity & Uniform-PAC across tabular, structural measures, FA, RFE, offline, POMDP \\
\textbf{Time window} & Pre-2010 & Pre-2009 & 2018--2025 \\
\textbf{Primary axis} & Algorithms and theory & PAC-MDP sample complexity & CSO: Coverage $\times$ Structure $\times$ Objective \\
\textbf{Guarantee types} & PAC/Regret (primarily tabular) & PAC-MDP bounds for finite state-action spaces & Uniform-PAC, instance-dependent, offline/pessimism, PAC-Bayes \\
\textbf{Practitioner tools} & Conceptual algorithms & Theoretical foundations & Decision tree; coverage gates; certificates; rate tables \\
\bottomrule
\end{tabular}
\caption{\textbf{Comparison with foundational surveys.} We build on Szepesvári's algorithmic treatment and Strehl et al.'s PAC-MDP framework, adding 2018--2025 results under a unified CSO lens with deployment-oriented tools.}
\label{tab:compare-surveys}
\end{table}

\section{The Coverage-Structure-Objective Framework}
\label{sec:cso}

\subsection{Why a unifying lens is needed}

The PAC RL literature from 2018 to 2025 spans a remarkable variety of settings: tabular and continuous state spaces, online and offline data access, worst-case and instance-dependent objectives, value-based and model-based algorithms.
Each paper introduces its own notation, assumptions, and complexity measures, making it difficult for a reader to compare results across settings or to determine which guarantee applies to a given problem.

Existing surveys organize results by algorithm family (model-based vs.\ model-free) or by setting (tabular vs.\ function approximation).
These axes are natural but insufficient: they do not explain why sample complexity changes when the data source shifts from online to offline, or why the same structural parameter (say, feature dimension $d$) enters differently in reward-free exploration than in pessimistic control.
A reader who wants to know ``I have an offline dataset with linear features and need an $\varepsilon$-optimal policy; what is the best available rate?'' must hunt through papers with incompatible notation to piece together the answer.

This survey proposes an alternative organizing principle that we call the \textbf{Coverage-Structure-Objective (CSO) framework}.
The CSO framework is not a theorem and does not appear in any single prior paper.
It is this survey's contribution as a reading lens: an observation about the recurring multiplicative structure of PAC bounds across the 2018--2025 literature, elevated to a deliberate organizational tool.

\subsection{The three axes and the generic rate}

We observe that nearly every PAC sample complexity bound surveyed in \S\ref{sec:tabular} through \S\ref{sec:pacbayes} can be parsed as a product of three interpretable factors, up to polylogarithmic terms and horizon polynomials:
\begin{equation}
\label{eq:cso}
N(\varepsilon,\delta) \;\approx\; \underbrace{\mathsf{Cov}}_{\text{coverage}} \;\times\; \underbrace{\mathsf{Comp}}_{\text{structure}} \;\times\; \mathrm{poly}(H) \;\times\; \varepsilon^{-2} \;\times\; \log(1/\delta).
\end{equation}
The symbol $\approx$ here means equality up to polylogarithmic factors in $(S, A, H, 1/\varepsilon, 1/\delta)$ and up to the precise polynomial degree in $H$, which varies by setting.
When we write $\mathrm{poly}(H)$, we mean a specific power of the horizon (for instance $H^3$ in the tabular case, $H^4$ for linear MDPs) that the template deliberately leaves abstract so that the three-factor decomposition remains visible.
The exact exponents are stated in the relevant sections and summarized in Table~\ref{tab:taxonomy}.

We now explain what each factor captures.

\paragraph{Coverage ($\mathsf{Cov}$)} reflects how data are obtained and how well they support the target policy.
In online settings, the agent creates its own coverage through exploration, so $\mathsf{Cov} = 1$: coverage is not a bottleneck.
The generative model setting, where the learner can query any state-action pair on demand, also gives $\mathsf{Cov} = 1$.
In offline RL, coverage is inherited from whatever behavior policy generated the dataset and is quantified by the concentrability coefficient $C_\star$ (Definition~\ref{def:concentrability}).
When $C_\star$ is large, the dataset provides little information about states that matter for the optimal policy, and $\mathsf{Cov} = \mathrm{poly}(C_\star)$ can dominate the entire sample complexity expression.
In reward-free exploration, coverage is neither given nor free but \emph{constructed} as a reusable resource: the exploration budget creates broad support for all possible downstream rewards, and the extra factor of $S$ in the tabular RFE bound (Theorem~\ref{thm:rfe-tab-dep}) reflects this investment.

\paragraph{Structure ($\mathsf{Comp}$)} measures the intrinsic complexity of the MDP or the function class used to approximate values or models.
In the tabular case, $\mathsf{Comp} = SA$, the number of free parameters in the transition and reward functions.
With linear features of dimension $d$, $\mathsf{Comp} = \mathrm{poly}(d)$, typically $d^3$ for standard LSVI-UCB bounds.
With kernel or RKHS value classes, $\mathsf{Comp} = d_{\mathrm{eff}}(\lambda)$, the effective dimension at regularization level $\lambda$, which is controlled by the spectral decay of the kernel integral operator.
With low-rank transitions, $\mathsf{Comp} = \mathrm{poly}(r)$ where $r$ is the factorization rank.
The structural complexity measures surveyed in \S\ref{sec:structure} (Bellman rank $B$, witness rank $W$, Bellman-Eluder dimension $d_{\mathrm{BE}}$, bilinear rank) are all instantiations of this axis, forming a hierarchy from tight-but-narrow (tabular) to broad-but-loose (finite $d_{\mathrm{BE}}$).

\paragraph{Objective ($\mathsf{Obj}$)} specifies what the learner must achieve.
The baseline is $(\varepsilon,\delta)$-PAC control: output an $\varepsilon$-optimal policy with probability $1 - \delta$.
Uniform-PAC strengthens this to simultaneous guarantees for all $\varepsilon$, which implies high-probability regret (Theorem~\ref{thm:uniformpac-regret}).
Instance-dependent identification replaces the worst-case structural factor with gap-weighted sums that can be much smaller (\S\ref{sec:tabular}).
Off-policy evaluation asks for accurate value estimates rather than optimal policies.
Each objective variation modifies the $\varepsilon^{-2}$ core or the horizon polynomial, but the product structure is preserved.

The value of this decomposition is organizational rather than mathematical.
It tells the reader which knob to turn when a bound is too large: reduce the structural term by using better features, improve coverage by collecting more diverse data, or relax the objective from control to evaluation.
And it makes cross-setting comparison immediate, because two results that differ only in one CSO coordinate can be compared by reading off the corresponding factor.

\subsection{Four examples that show the template in action}

To make the template concrete, we instantiate it for four canonical results that appear later in the survey.
The progression illustrates how changing one CSO coordinate at a time alters the sample complexity in predictable ways.

\paragraph{Tabular online PAC (\S\ref{sec:tabular}).}
With online access, the agent explores freely, so $\mathsf{Cov} = 1$.
The structural complexity of a finite MDP is $\mathsf{Comp} = SA$, and the horizon enters as $H^3$.
The resulting rate is $N = \tTheta(SAH^3/\varepsilon^2)$, confirmed as minimax-optimal by Zhang et al.~\citep{zhang2024settling}.
This is the simplest CSO cell and the calibration point for all other results.

\paragraph{Linear MDP with LSVI-UCB (\S\ref{sec:function}).}
Keeping online access ($\mathsf{Cov} = 1$) but replacing the tabular structure with linear features, the structural term becomes $\mathsf{Comp} = d^3$ and the horizon exponent rises to $H^4$.
The rate is $N = \tO(d^3 H^4/\varepsilon^2)$~\citep{jin2020linear}.
The increase from $H^3$ to $H^4$ is a genuine cost of function approximation: estimation errors are correlated across states sharing features, and these correlations compound through Bellman backups.
Compared to the tabular case, we have traded the $SA$ factor for $d^3$, which is a large gain when $d \ll SA$.

\paragraph{Offline linear MDP with pessimism (\S\ref{sec:offline}).}
Now change the access mode from online to offline, keeping the linear structure.
The coverage axis shifts from $\mathsf{Cov} = 1$ to $\mathsf{Cov} = \mathrm{poly}(C_\star)$, where $C_\star$ measures how much the optimal policy's state visitation deviates from the data distribution.
The rate becomes $n = \tO(\mathrm{poly}(d, C_\star, H)/\varepsilon^2)$~\citep{jin2021pevi,shi2022pql}.
Coverage is now the bottleneck: even with $d = 5$, if $C_\star = 10^4$ the bound is vacuous, and no amount of structural simplicity can compensate.
This example shows why coverage diagnostics (\S\ref{subsec:coverage}) are essential before deploying offline policies.

\paragraph{Reward-free exploration, tabular (\S\ref{sec:rfe}).}
Return to online access and tabular structure, but change the objective: the agent must build a dataset that supports $\varepsilon$-optimal planning for \emph{any} reward function, not just one.
This strengthened objective inflates the coverage axis to $\mathsf{Cov} = S$ (the cost of covering occupancies for all possible downstream rewards), giving $N = \tO(S^2 A\,\mathrm{poly}(H)/\varepsilon^2)$~\citep{jin2020rewardfree}.
The extra factor of $S$ relative to standard online learning is a coverage investment that amortizes across arbitrarily many downstream tasks; in a multi-reward pipeline, this upfront cost can be far cheaper than running separate exploration for each reward.

\subsection{Benefits and limitations of the CSO lens}

We claim three benefits for the CSO decomposition.

The first is \emph{navigability}.
A reader can locate any result in the survey by specifying its CSO coordinates: identify the access mode (online, generative, offline, reward-free) to determine coverage, characterize the function class (tabular, linear, kernel, low-rank, general) to determine structure, and fix the learning goal (PAC control, uniform-PAC, identification, evaluation) to determine the objective.
Table~\ref{tab:taxonomy} cross-references every major result by these coordinates, and the recommended workflow in \S\ref{subsec:cso-workflow} walks through this process step by step.

The second is \emph{diagnosis}.
When a guarantee is vacuous, the CSO parsing identifies which axis is responsible and suggests remedies.
If the structural term dominates, invest in better features or a richer function class.
If coverage dominates, collect more diverse data or relax the objective to evaluation rather than control.
If the horizon polynomial is the problem, consider state abstraction or temporal hierarchy to reduce the effective horizon.

The third is \emph{frontier identification}.
The hardest open problems in \S\ref{sec:open} are precisely those where multiple axes interact.
Offline RL with misspecified features couples coverage and structure: neither improving features alone nor improving data alone resolves the problem.
Agnostic low-rank learning couples structure and computation: the statistical rates are known, but polynomial-time algorithms require additional assumptions.
Instance-dependent identification with function approximation couples structure and objective: the gap-weighted rates from tabular settings do not transfer without new tools for measuring reachability in high-dimensional spaces.

The framework also has clear limitations that we flag here to avoid overclaiming.
It does not capture computational complexity: two settings with the same CSO coordinates may differ dramatically in whether a polynomial-time algorithm exists.
It abstracts away horizon exponents that differ meaningfully across settings ($H^3$ vs.\ $H^4$ vs.\ $H^6$ can matter in practice).
It does not apply to non-PAC objectives such as Bayesian regret or simple regret in pure exploration without the PAC wrapper.
And the multiplicative decomposition is approximate; in some settings, coverage and structure interact through products of logarithms or through shared parameters that do not factor cleanly.
We flag these limitations where they matter in later sections and provide exact exponents alongside the CSO parsing.

\subsection{Using the framework: a workflow}
\label{subsec:cso-workflow}

When confronting a new RL problem, we recommend the following sequence.

Start by identifying the access mode.
Can the agent interact with the environment online, or is there only a fixed dataset?
Is a generative model (simulator) available?
Will the reward be specified later, suggesting a reward-free setup?
The answer determines the coverage axis: online and generative access give $\mathsf{Cov} = 1$; offline access introduces $\mathsf{Cov} = \mathrm{poly}(C_\star)$; reward-free exploration has an inflated coverage cost that amortizes across tasks.

Next, characterize the function class.
Is the problem tabular (small, finite state and action spaces)?
Are features available, and if so, are they linear, kernelized, or learned?
Does the environment have latent low-rank structure?
The answer determines the structural axis and points to the relevant section of the survey.

Then fix the objective.
Is the goal to find a single near-optimal policy (PAC control), to guarantee good performance at all accuracy levels (uniform-PAC), to identify the best policy with instance-dependent sample complexity, or to evaluate a given policy without improvement?

With these three coordinates in hand, read off the rate from Table~\ref{tab:taxonomy} and the corresponding section.
Before invoking the guarantee, verify the underlying assumptions using the diagnostics of \S\ref{sec:practice}: check realizability via Bellman residual tests (Algorithm~\ref{alg:bellman}), estimate coverage via density ratios and leverage scores (Algorithm~\ref{alg:coveragegate}), and gate deployment on policy certificates (Definition~\ref{def:cert}).

The rest of the survey instantiates this workflow for each CSO cell, beginning with the formal definitions in \S\ref{sec:preliminaries}.


\begin{table}[t]
\centering
\small
\setlength{\tabcolsep}{4pt}
\begin{tabular}{llllll}
\toprule
\textbf{Setting} & $\mathsf{Cov}$ & $\mathsf{Comp}$ & $\mathrm{poly}(H)$ & \textbf{Rate} & \textbf{Section} \\
\midrule
Tabular online & $1$ & $SA$ & $H^3$ & $\tTheta(SAH^3/\varepsilon^2)$ & \S\ref{sec:tabular} \\
Tabular BPI & $1$ & $\sum \frac{q_h}{\Delta^2}$ & varies & instance-dep. & \S\ref{sec:tabular} \\
Linear MDP & $1$ & $d^3$ & $H^4$ & $\tO(d^3 H^4/\varepsilon^2)$ & \S\ref{sec:function} \\
Kernel/RKHS & $1$ & $d_{\mathrm{eff}}(\lambda)$ & $H^4$--$H^6$ & $\tO(d_{\mathrm{eff}} \cdot \mathrm{poly}(H)/\varepsilon^2)$ & \S\ref{sec:function} \\
Low-rank MDP & $1$ & $\mathrm{poly}(r)$ & varies & $\tO(\mathrm{poly}(r,H)/\varepsilon^2)$ & \S\ref{sec:rich} \\
Block MDP & $1$ & $\mathrm{poly}(m) \cdot A$ & varies & $\tO(\mathrm{poly}(m)AH^{\cdot}/\varepsilon^2)$ & \S\ref{sec:rich} \\
Bellman rank $B$ & $1$ & $\mathrm{poly}(B)$ & varies & $\tO(\mathrm{poly}(B,H)/\varepsilon^2)$ & \S\ref{sec:structure} \\
RFE (tabular) & $S$ & $SA$ & $\mathrm{poly}(H)$ & $\tO(S^2A\,\mathrm{poly}(H)/\varepsilon^2)$ & \S\ref{sec:rfe} \\
RFE (linear) & $1$ & $\mathrm{poly}(d)$ & varies & $\tO(\mathrm{poly}(d,H)/\varepsilon^2)$ & \S\ref{sec:rfe} \\
Offline linear & $\mathrm{poly}(C_\star)$ & $\mathrm{poly}(d)$ & varies & $\tO(\mathrm{poly}(d,C_\star,H)/\varepsilon^2)$ & \S\ref{sec:offline} \\
Offline tabular & $\mathrm{poly}(C_\star)$ & $SA$ & varies & minimax-optimal & \S\ref{sec:offline} \\
\bottomrule
\end{tabular}
\caption{\textbf{CSO coordinates of major results.} Each row decomposes a PAC bound into Coverage ($\mathsf{Cov}$), Structure ($\mathsf{Comp}$), and horizon scaling. ``Varies'' indicates setting-dependent exponents stated in the referenced section. All rates suppress $\log(1/\delta)$ and polylogarithmic factors.}
\label{tab:taxonomy}
\end{table}

\section{Preliminaries: PAC Learning Meets Sequential Decision-Making}
\label{sec:preliminaries}

The notion of Probably Approximately Correct (PAC) learning originated in computational learning theory with Valiant's framework for supervised classification~\citep{valiant1984pac}.
A PAC learner guarantees that, with probability at least $1-\delta$, the hypothesis it outputs has error at most $\varepsilon$, after observing a number of examples polynomial in $1/\varepsilon$, $1/\delta$, and problem-dependent parameters.
Transferring this idea to sequential decision-making requires confronting three challenges absent from the supervised setting: the agent's actions influence the data it observes, credit for a good outcome must be assigned across a sequence of decisions, and the distribution over states shifts as the policy changes.

The formal apparatus for sequential decisions rests on the theory of Markov decision processes (MDPs), developed by Bellman~\citep{bellman1957dynamic} and consolidated by Howard~\citep{howard1960dynamic} and Puterman~\citep{puterman1994mdp}.
We adopt the episodic finite-horizon MDP as our canonical model, following Sutton and Barto~\citep{suttonbarto2018book} for algorithmic intuition and Szepesvári~\citep{szepesvari2010algorithms} for analytical foundations.

This section fixes notation, defines the PAC and uniform-PAC criteria that underpin the entire survey, introduces the structural and coverage assumptions that later sections instantiate, and states baseline results that calibrate what is achievable.
Readers already comfortable with episodic MDPs and PAC definitions may wish to skim through to the baseline results in \S\ref{subsec:baselines} and the notation summary in \S\ref{subsec:shorthand}, returning to earlier subsections as needed.

\subsection{The episodic MDP model}

An episodic finite-horizon MDP is a tuple $M = (\cS, \cA, \{P_h\}_{h=1}^H, \{r_h\}_{h=1}^H, H, \rho)$.
The state space $\cS$ has cardinality $|\cS| = S$, and the action space $\cA$ has cardinality $|\cA| = A$.
At each stage $h \in \{1, \ldots, H\}$, the agent observes a state $s_h \in \cS$, selects an action $a_h \in \cA$, receives reward $r_h(s_h, a_h) \in [0,1]$, and transitions to a new state $s_{h+1} \sim P_h(\cdot \mid s_h, a_h)$.
The initial state is drawn from $\rho$, and the episode terminates after stage $H$.

A policy $\pi = (\pi_1, \ldots, \pi_H)$ maps states to distributions over actions at each stage; it may be nonstationary (stage-dependent), which is necessary for optimality in the finite-horizon setting.
The value function $V_h^\pi(s) = \E^\pi[\sum_{t=h}^H r_t(s_t, a_t) \mid s_h = s]$ measures the expected cumulative reward from state $s$ at stage $h$ onward under policy $\pi$, and the action-value function $Q_h^\pi(s,a) = \E^\pi[\sum_{t=h}^H r_t(s_t, a_t) \mid s_h = s, a_h = a]$ conditions additionally on the first action.
These functions satisfy the Bellman equations introduced by Bellman~\citep{bellman1957dynamic}: $Q_h^\pi(s,a) = r_h(s,a) + \E_{s' \sim P_h(\cdot|s,a)}[V_{h+1}^\pi(s')]$ with $V_h^\pi(s) = \E_{a \sim \pi_h(\cdot|s)}[Q_h^\pi(s,a)]$.

The optimal value $V^\star_h(s) = \sup_\pi V_h^\pi(s)$ and optimal action-value $Q^\star_h(s,a) = \sup_\pi Q_h^\pi(s,a)$ are attained by a deterministic optimal policy $\pi^\star$ that exists in every finite MDP~\citep{puterman1994mdp}.
The Bellman optimality operator at stage $h$ acts on functions $f_{h+1} : \cS \times \cA \to \R$ via
\[
(\mathcal{T}_h f)(s,a) \;=\; r_h(s,a) + \E_{s' \sim P_h(\cdot \mid s,a)}\!\big[\max_{a'} f_{h+1}(s', a')\big],
\]
so that $Q_h^\star = \mathcal{T}_h Q_{h+1}^\star$ for all $h$.
The Bellman error of a candidate function $f_h$ at a state-action pair $(s,a)$ measures how far $f_h$ is from satisfying this fixed-point equation.
Controlling Bellman errors is a recurring motif throughout the survey: in online settings, optimistic bonuses inflate estimates to encourage exploration; in offline settings, pessimistic penalties deflate estimates to guard against poor coverage; and in both cases, the sample complexity is governed by how quickly Bellman errors concentrate around zero.

Discounted infinite-horizon problems map into this framework by setting an effective horizon $H_{\mathrm{eff}} \asymp (1-\gamma)^{-1}$ where $\gamma$ is the discount factor.
We state all results for the finite-horizon case and note conversions where they apply.

\subsection{PAC and uniform-PAC guarantees}

The central quality criterion in this survey is the fixed-confidence (PAC) guarantee.
In the supervised setting, PAC learning requires a hypothesis that is $\varepsilon$-close to optimal with high probability after a polynomial number of samples~\citep{valiant1984pac}.
In RL, the analogous requirement asks for an $\varepsilon$-optimal policy after a controlled number of episodes of interaction.

\begin{definition}[$(\varepsilon,\delta)$-PAC (fixed-confidence control)]
\label{def:pac-prelim}
An algorithm $\mathsf{Alg}$ is $(\varepsilon,\delta)$-PAC if, with probability at least $1-\delta$, after at most $N(\varepsilon,\delta)$ episodes it outputs a policy $\hat\pi$ satisfying $V_1^\star(\rho) - V_1^{\hat\pi}(\rho) \le \varepsilon$.
\end{definition}

This definition fixes a single accuracy target $\varepsilon$ in advance.
In many applications, however, the acceptable error tolerance is not known at the outset, or we wish to derive regret bounds from PAC-style results.
Dann, Lattimore, and Brunskill~\citep{uniformpac2017} introduced uniform-PAC to address this need, strengthening the guarantee to hold across all accuracy levels at once.

\begin{definition}[Uniform-PAC]
\label{def:uniform-pac}
An algorithm is uniform-PAC if, with probability at least $1-\delta$, for all $\varepsilon > 0$ simultaneously, the number of episodes in which the algorithm plays an $\varepsilon$-suboptimal policy is at most $N(\varepsilon,\delta)$.
\end{definition}

The distinction matters.
A single-$\varepsilon$ PAC algorithm can behave arbitrarily at accuracy levels other than the one it was tuned for; a uniform-PAC algorithm must perform well everywhere.
This extra strength buys a direct bridge to cumulative regret, which we state here and reference throughout the survey.

\begin{theorem}[Uniform-PAC implies high-probability regret]
\label{thm:uniformpac-regret}
If an algorithm is uniform-PAC with budget $N(\varepsilon,\delta)$, then with probability $1-\delta$ its cumulative regret after $K$ episodes satisfies
\[
\mathrm{Regret}(K) \;=\; \cO\!\bigg(\int_0^H \min\{K,\, N(\varepsilon,\delta)\}\, d\varepsilon\bigg).
\]
When $N(\varepsilon,\delta)$ has polynomial dependence on $(S,A,H,1/\varepsilon)$, this recovers near-minimax tabular regret rates~\citep{uniformpac2017}.
\end{theorem}

The conversion works by observing that the regret contribution from accuracy level $\varepsilon$ is at most $\varepsilon \cdot N(\varepsilon,\delta)$ (each of the $N(\varepsilon,\delta)$ bad episodes costs at most $\varepsilon$), and integrating over all accuracy levels yields the stated bound.
This is tight: uniform-PAC is strictly stronger than single-$\varepsilon$ PAC, and the regret integral cannot be improved without additional assumptions~\citep{uniformpac2017}.
For the rest of the survey, any uniform-PAC result automatically yields a regret bound via this theorem, linking the fixed-confidence and online-learning perspectives without further work.

\subsection{Function classes, realizability, and completeness}

Beyond the tabular regime, where every state-action pair can be visited and estimated separately, sample complexity depends on the richness of the function class used to approximate values or models.
Three concepts recur throughout the structured settings of \S\ref{sec:structure} through \S\ref{sec:rich}.

\begin{definition}[Function class and realizability]
\label{def:function-class}
Let $\cF$ be a class of candidate value or $Q$-functions.
Realizability asserts that $Q^\star \in \cF$ (or $V^\star \in \cF$): the true optimal value function belongs to the hypothesis class.
In the agnostic setting, $Q^\star$ need not belong to $\cF$, and performance is measured relative to $\inf_{f \in \cF} \|f - Q^\star\|$, the best approximation error within the class.
\end{definition}

Realizability is a strong assumption: it says the modeler chose the right function class.
When it holds, algorithms can focus on statistical estimation (narrowing a confidence set that contains $Q^\star$); when it fails, approximation error compounds through Bellman backups and can overwhelm the statistical gains from function approximation.
Checking realizability in practice is the subject of the misspecification diagnostic in \S\ref{subsec:misspec}.

\begin{definition}[Bellman completeness]
\label{def:bellman-completeness}
A function class $\cF$ is Bellman-complete if applying the Bellman operator to any member of $\cF$ yields another member: $\mathcal{T}_h \cF \subseteq \cF$ for all $h$.
\end{definition}

Bellman completeness is a closure condition that prevents error amplification through backward induction.
If $\cF$ contains $Q^\star_H$ (the last stage) and is closed under $\mathcal{T}_h$, then backward induction starting from stage $H$ stays within $\cF$ at every stage, and standard regression-based algorithms converge.
Without completeness, the Bellman backup of a function in $\cF$ may lie outside $\cF$, forcing the algorithm to project back into $\cF$ at each stage and introducing projection errors that accumulate over the horizon.
This accumulation is the single most common failure mode in function-approximation RL: the algorithm appears to converge but settles on a suboptimal fixed point because compounding projection errors steer it away from $Q^\star$.
Bellman completeness is rarely verifiable in advance for a given function class and MDP, which is why the residual diagnostic of \S\ref{subsec:misspec} is a practical necessity rather than a theoretical nicety.

\begin{definition}[Covering number]
\label{def:covering}
The covering number $\mathcal{N}(\epsilon, \cF, \|\cdot\|)$ is the size of the smallest $\epsilon$-net of $\cF$ under norm $\|\cdot\|$: the minimum number of balls of radius $\epsilon$ needed to cover $\cF$.
\end{definition}

Covering numbers appear in PAC analyses through union bounds: to guarantee that a confidence set contains $Q^\star$, the algorithm must control estimation error uniformly over $\cF$, and the cost of this uniform control scales logarithmically with $\mathcal{N}$.
For parametric classes (linear, low-rank), covering numbers grow polynomially in dimension and $1/\epsilon$; for nonparametric classes (kernel, neural), the growth depends on the spectral decay or architectural complexity, and the effective dimension $d_{\mathrm{eff}}(\lambda)$ of \S\ref{sec:function} provides a refined surrogate.

\subsection{Coverage, access models, and occupancy}

How data are collected, and how well they cover the state-action space relevant to the target policy, is the third axis of the CSO framework.
The formal objects are occupancy measures and concentrability coefficients.

For a policy $\pi$ and initial distribution $\rho$, the occupancy measure $d_h^\pi(s,a)$ is the probability of visiting state-action pair $(s,a)$ at stage $h$ when following $\pi$ from $\rho$.
In online RL, the agent controls which policies generate data and can steer occupancy toward informative regions through exploration bonuses.
In offline RL, the occupancy $\mu_h(s,a)$ of the behavior policy is fixed, and the agent must work with whatever support $\mu$ provides.

\begin{definition}[Concentrability coefficient]
\label{def:concentrability}
The concentrability coefficient measures how much the optimal policy's visitation deviates from the data distribution:
\[
C_\star \;=\; \max_{h \in [H]} \bigg\| \frac{d_h^{\pi^\star}}{\mu_h} \bigg\|_\infty.
\]
When $C_\star$ is small, the dataset has reasonable support everywhere the optimal policy visits, and offline learning is feasible.
When $C_\star$ is large, the dataset provides little information about states that matter for $\pi^\star$, and guarantees degrade or become vacuous.
When $C_\star$ is infinite (the behavior policy never visits some states that $\pi^\star$ does), consistent policy improvement is impossible from the data alone.
\end{definition}

The concentrability coefficient is defined in terms of the unknown $\pi^\star$, which makes it impossible to compute exactly from data.
In practice, surrogate quantities (density-ratio estimates, ridge leverage scores, effective sample sizes) provide approximate coverage diagnostics; \S\ref{subsec:coverage} develops concrete procedures for this estimation.

\begin{definition}[Access models]
\label{def:access}
Three access modes appear throughout the survey.
In \emph{online} interaction, the agent sequentially collects trajectories under its evolving policy, choosing actions and observing transitions in real time.
A \emph{generative model} allows i.i.d.\ sampling from $P_h(\cdot \mid s, a)$ for any query $(s, a, h)$; this is a stronger oracle than online access because it permits targeted queries without committing to a trajectory.
In the \emph{offline} setting, a fixed dataset $\cD$ of trajectories collected by an unknown behavior policy is given, and no further interaction with the environment is permitted.
\end{definition}

Online and generative-model access set $\mathsf{Cov} = 1$ in the CSO template because the agent can create its own coverage.
Offline access introduces $\mathsf{Cov} = \mathrm{poly}(C_\star)$: coverage is inherited, not constructed, and its quality directly governs the sample complexity.
Reward-free exploration (\S\ref{sec:rfe}) occupies an intermediate position: the agent has online access but must build coverage sufficient for all possible rewards, paying an extra cost that appears in the coverage term.

In rich-observation settings (\S\ref{sec:structure}, \S\ref{sec:rich}), observations $x \in \cX$ may be high-dimensional (images, sensor streams) even when the underlying dynamics are governed by a small latent state $z \in [m]$ with $m \ll |\cX|$.
The gap between observation complexity and latent complexity is what makes representation learning both necessary and possible in these settings.

\subsection{Baseline results}
\label{subsec:baselines}

With notation and definitions in place, we state the baselines that calibrate all subsequent results.
These are the sharpest known guarantees in the simplest (tabular) setting; later sections show how structure reduces the $(S,A)$ factors but preserves the $\varepsilon^{-2}$ and $\mathrm{poly}(H)$ scalings.

\begin{theorem}[Tabular minimax sample complexity]
\label{thm:tabular-minimax}
In finite-horizon tabular MDPs, the minimax sample complexity for $(\varepsilon,\delta)$-PAC learning is
\[
N(\varepsilon,\delta) \;=\; \tTheta\!\bigg(\frac{SAH^3}{\varepsilon^2}\bigg),
\]
and the minimax regret after $K$ episodes is $\tTheta(\sqrt{SAH^3 K})$, both up to polylogarithmic factors.
The upper bound is achieved by optimistic value-iteration variants such as UCB-VI~\citep{azar2017ucbvi}; the lower bound shows that the $SAH^3$ product is information-theoretically necessary~\citep{domingues2021lowerbounds,zhang2024settling}.
\end{theorem}

The $H^3$ exponent is the tightest among all settings in the survey and arises from three independent contributions of the horizon: the length of the episode scales the total variance of returns, variance accumulates across stages, and a stage-wise union bound contributes a further factor.
Zhang et al.~\citep{zhang2024settling} settled the optimal online sample complexity, confirming $H^3$ as the correct exponent, and Domingues et al.~\citep{domingues2021lowerbounds} provided refined lower bounds pinning down each parameter's contribution.

The remaining baseline results (reward-free exploration, linear MDPs, Bellman-Eluder dimension, offline RL) were previewed in the introduction (\S\ref{sec:introduction}) and are developed in full in their respective sections.
Rather than restating them here, we note the pattern they share: each factors into the product form of Equation~\eqref{eq:cso}, with structural parameters replacing $SA$, coverage terms entering for offline or reward-free settings, and $\varepsilon^{-2}\log(1/\delta)$ as the irreducible statistical core.
Understanding when and why each factor can be reduced is the subject of the rest of this survey.

\subsection{Notation summary}
\label{subsec:shorthand}

We write $\Phi_{\mathrm{tab}}(S,A,H) \equiv SAH^3$ for the canonical tabular factor.
The notation $\tO(\cdot)$ suppresses polylogarithmic factors in $(S, A, H, 1/\varepsilon, 1/\delta)$, and $\tTheta(\cdot)$ and $\tOmega(\cdot)$ denote the corresponding tight and lower-bound variants.
When we say ``polynomial dependence,'' we mean polynomial in the stated parameters with degree that is bounded but not always explicitly tracked; when exact exponents are known, we state them or point to the primary source.
A complete notation table, consolidating all symbols used in the survey, appears in Appendix~\ref{app:notation}.


\section{Tabular Baselines: Minimax and Instance-Dependent Guarantees}
\label{sec:tabular}

The tabular setting, where both the state space and action space are finite and small enough to enumerate, is where the sharpest PAC and uniform-PAC guarantees are known.
No function approximation is involved: the algorithm can maintain explicit estimates for every state-action pair at every stage.
Every result in later sections must recover tabular rates as a special case (e.g., linear MDPs with one-hot features at $d = SA$), making these bounds the universal calibration point.
The central questions are: what is the minimax sample complexity to learn an $\varepsilon$-optimal policy with confidence $1-\delta$, and when can instance-dependent structure yield substantially faster identification?

\subsection{The minimax picture}

Theorem~\ref{thm:tabular-minimax} established the minimax sample complexity as $\tTheta(SAH^3/\varepsilon^2)$.
This result was built over a sequence of increasingly tight analyses.
Early PAC-MDP bounds by Strehl, Li, and Littman~\citep{strehl2009pacjmlr} established polynomial sample complexity in finite MDPs, with subsequent work by Azar, Osband, and Munos~\citep{azar2017ucbvi} introducing UCB-VI with $\sqrt{H}$-type regret and by Jaksch, Ortner, and Auer~\citep{jaksch2010ucrl2} addressing the average-reward case.
The modern picture crystallized with Domingues et al.~\citep{domingues2021lowerbounds}, who derived refined lower and upper bounds for episodic tabular RL, and Zhang et al.~\citep{zhang2024settling}, who settled the optimal dependence on all parameters.

The $H^3$ exponent deserves comment, as it is the tightest among all settings in the survey.
It arises from three independent contributions of the horizon: the length of the episode scales the total variance of returns, variance accumulates across stages, and the cost of a union bound over $H$ stages adds a further factor.
In structured settings (\S\ref{sec:function}), these contributions become entangled through correlated estimation errors, yielding higher exponents ($H^4$ for linear MDPs, up to $H^6$ for kernel models).

The uniform-PAC framework connects these fixed-confidence results directly to regret.
By Theorem~\ref{thm:uniformpac-regret}, any uniform-PAC algorithm with budget $N(\varepsilon,\delta) = \tO(SAH^3/\varepsilon^2)$ automatically achieves high-probability regret $\tO(\sqrt{SAH^3 K})$, which is minimax-optimal up to logarithms~\citep{uniformpac2017}.
This bridge means that PAC and regret analyses are not competing frameworks but complementary views of the same statistical challenge.

\subsection{Instance-dependent identification}

Minimax rates are pessimistic: they reflect the hardest possible instance.
When the MDP has well-separated optimal actions on reachable states, identification can be dramatically cheaper.

\begin{definition}[Best-policy identification, gaps, and reachability]
\label{def:bpi}
For each stage $h$ and state $s$, the action gap $\Delta_h(s,a) = Q_h^\star(s, a_h^\star(s)) - Q_h^\star(s,a)$ measures how suboptimal action $a$ is.
The reachability factor $q_h(s) = \sup_\pi d_h^\pi(s)$ captures how easily state $s$ can be reached at stage $h$.
Best-policy identification (BPI) asks for algorithms whose sample complexity scales with these instance-dependent quantities rather than with $(S,A)$ alone.
\end{definition}

The intuition is straightforward.
If the optimal action at every reachable state is separated from the alternatives by a large gap $\Delta$, then the algorithm does not need to resolve fine-grained differences between similar actions; it only needs enough data to confirm which action is best.
Conversely, if a state is unreachable (or nearly so) under every policy, wasting samples trying to identify the optimal action there does not help.
The instance-dependent rate captures both effects: it weights the difficulty of each state-action pair by how reachable the state is and how hard the action is to distinguish from optimal.

\begin{theorem}[Instance-dependent BPI]
\label{thm:tab-id}
There exist optimistic identification algorithms achieving
\[
N_{\mathrm{BPI}}(\varepsilon,\delta) \;=\; \tO\!\bigg(\sum_{h=1}^H \sum_{s \in \cS} q_h(s) \sum_{\substack{a \in \cA:\\ \Delta_h(s,a) > \varepsilon}} \frac{1}{\Delta_h(s,a)^2}\bigg),
\]
and matching lower bounds show that the dependence on inverse squared gaps and reachability is unavoidable up to logarithms.
\end{theorem}

Wagenmaker, Simchowitz, and Jamieson~\citep{wagenmaker2022idpac} established these rates, revealing regimes where identification is far easier than worst-case learning.
When large gaps exist on all reachable states, $N_{\mathrm{BPI}}$ can be as small as $\tO(H^2 / \Delta_{\min}^2)$ regardless of $(S,A)$, because the algorithm only needs to confirm the obvious choice at each state.
When gaps vary widely across states, the sum concentrates on the few ``hard'' states where actions are nearly tied, and the rest contribute negligibly.
Tirinzoni, Al-Marjani, and Kaufmann~\citep{tirinzoni2023optimistic} refined the analysis with optimistic algorithms and clarified how reachability and gap structure interact: a state with a tiny gap contributes to the sample complexity only in proportion to how often any policy visits it.

Through the CSO lens, instance-dependent identification replaces the worst-case structural factor $\mathsf{Comp} = SA$ with a gap-weighted sum that can be much smaller.
The coverage term remains $\mathsf{Cov} = 1$ (online access), and the objective shifts from uniform-PAC to targeted identification.

\subsection{Policy certificates}

A complementary perspective on accountability comes from policy certificates: data-dependent, per-episode bounds on suboptimality.

\begin{definition}[Policy certificate]
\label{def:cert}
A policy certificate at episode $t$ is a quantity $U_t$, computed from the data collected so far, such that with probability at least $1-\delta$, $V_1^\star(\rho) - V_1^{\pi_t}(\rho) \le U_t$.
Certificates provide auditable guarantees without waiting for the algorithm to terminate.
\end{definition}

The value of certificates is that they transform a retrospective guarantee (``after $N$ episodes, the final policy is good'') into a prospective one (``right now, at episode $t$, the current policy is at most $U_t$-suboptimal'').
Dann et al.~\citep{dann2019certificates} showed that certificates can be computed efficiently in tabular MDPs and that their cumulative sum tracks regret up to lower-order terms.
This makes certificates a natural deployment gate: a practitioner running an RL pipeline can monitor $U_t$ in real time and deploy the current policy only when $U_t$ falls below the desired tolerance, or collect more data if the certificate remains large.
The certificate approach is especially attractive in safety-sensitive settings where a post hoc ``the algorithm converged'' claim is insufficient and auditors require per-decision accountability.

\subsection{Practical considerations}

In tabular problems, practitioners should prefer algorithms with explicit confidence sets (optimistic bonuses or posterior sampling) that ensure uniform-PAC behavior, rather than heuristics without quantifiable uncertainty.
When identification is the primary goal, exploration should prioritize states with high estimated reachability and ambiguous gaps; spending samples on state-action pairs where the gap $\Delta_h(s,a)$ is already confidently large does not improve the output policy and slows convergence.

All tabular PAC claims are conditional on the tabular assumption itself.
When $S$ or $A$ grow too large for explicit enumeration, rates revert to those of function approximation (\S\ref{sec:function}), and the guarantees of this section serve only as calibration points: any structured bound must recover $\tTheta(SAH^3/\varepsilon^2)$ when the structure degenerates to tabular.

\paragraph{Diagnosing when tabular is insufficient: a worked example.}
Consider CartPole with state $(x, \dot{x}, \theta, \dot{\theta})$ and actions $\{\text{left}, \text{right}\}$.
With continuous states, $S = \infty$ and tabular methods cannot be applied.
One might discretize each dimension into bins, but the resulting $S$ grows exponentially in the state dimension: even 10 bins per dimension gives $S = 10^4$, and the tabular sample complexity $\tTheta(10^4 \cdot 2 \cdot H^3 / \varepsilon^2)$ becomes prohibitive for moderate $H$.
A better approach is to posit linear features $\phi(s,a) = [x, \dot{x}, \theta, \dot{\theta}, \mathbf{1}\{a = \text{right}\}]^\top$, reducing the problem to $d = 5$.
To verify whether this linear structure is adequate, collect transitions under a random policy, fit $\hat{Q}_h = \phi^\top w_h$ via ridge regression at each stage, and compute held-out Bellman residuals.
If residuals remain small and stable across stages, the linear class is adequate and the bounds of \S\ref{sec:function} apply with $d = 5$.
If residuals grow with $h$ or show systematic patterns (e.g., large errors near $\theta \approx \pm \pi/2$ where the dynamics are nonlinear), the linear class is misspecified.
In that case, the practitioner should enrich the features (e.g., add polynomial terms or trigonometric functions of $\theta$), switch to kernel methods (\S\ref{sec:function}), or investigate low-rank structure (\S\ref{sec:rich}).
The formal version of this diagnostic appears as Algorithm~\ref{alg:bellman} in \S\ref{subsec:misspec}.


\section{Structural Complexity: What Makes RL Learnable Beyond Tabular?}
\label{sec:structure}

When the state-action space is too large for tabular methods, the feasibility of PAC learning hinges on structural properties of the MDP and the function class used to approximate values or models.
Between 2017 and 2021, a family of complexity measures emerged that characterize when reinforcement learning with rich observations admits polynomial sample complexity.
These measures (Bellman rank, witness rank, the Bellman-Eluder dimension, and bilinear rank) replace the tabular parameters $(S,A)$ with problem-dependent quantities that can be much smaller, yielding guarantees that scale with the intrinsic difficulty of the problem rather than the size of the observation space.

This section traces the development of these measures, explains how they relate to each other and to tabular baselines, and discusses the computational assumptions (oracle efficiency) that separate statistical learnability from tractable algorithms.

\subsection{From tabular to structured: the need for complexity measures}

In tabular RL, sample complexity is governed by $SA$, the number of parameters needed to describe the transition and reward functions.
With function approximation, the relevant quantity is not the number of states but the capacity of the value or model class to represent distinct Bellman backup structures.
The challenge is to identify the right notion of capacity: covering numbers capture the size of the hypothesis space but do not account for the sequential structure of RL, where errors propagate backward through Bellman equations.

Consider a concrete example.
Suppose the state space has $10^6$ states, but the $Q$-function at every stage can be written as a linear combination of $d = 20$ known features.
The covering number of the linear class scales with $d$, not $S$, so a union-bound argument suggests sample complexity proportional to $d$ rather than $S$.
But this reasoning ignores Bellman structure: even if each stage's regression is easy, errors at stage $h+1$ feed into the regression target at stage $h$, and controlling this propagation requires more than just a small covering number.
The complexity measures in this section formalize precisely what additional structure is needed to tame this error propagation.

Jiang et al.~\citep{jiang2017olive} addressed this gap by introducing Bellman rank in the context of contextual decision processes (CDPs).
The key insight is that if the expected Bellman error of any candidate value function, evaluated under any admissible distribution, admits a low-rank factorization, then an elimination procedure can identify an $\varepsilon$-optimal policy with sample complexity polynomial in the rank and horizon.

\begin{definition}[Bellman rank~\citep{jiang2017olive}]
\label{def:bellman-rank}
A CDP has Bellman rank $B$ if, for each stage $h$, there exist feature maps $\phi_h : \cX \times \cA \to \R^B$ and $\psi_h : \cF \to \R^B$ (with bounded norms) such that the expected Bellman error of any $f \in \cF$ under any admissible distribution $\nu_h$ factors as
$\mathcal{E}_h(f; \nu_h) = \langle \E_{(x,a) \sim \nu_h} \phi_h(x,a),\, \psi_h(f) \rangle$.
\end{definition}

The factorization means that Bellman errors live in a $B$-dimensional space rather than the full function space.
An elimination algorithm can test candidate value functions against observed data and rule out those whose Bellman errors are inconsistent, narrowing the candidate set by at least one dimension per round.
After $\mathrm{poly}(B)$ rounds, only near-optimal candidates survive.

\begin{theorem}[Learning under Bellman rank]
\label{thm:olive}
Under realizability and Bellman rank $B$, elimination-style algorithms (e.g., OLIVE) achieve $N(\varepsilon,\delta) = \tO(\mathrm{poly}(B, H) \cdot \varepsilon^{-2} \cdot \log(1/\delta))$, using the function class $\cF$ for evaluation and greedy improvement~\citep{jiang2017olive}.
\end{theorem}

\subsection{Model-based identification: witness rank}

Bellman rank is a value-based notion: it characterizes the complexity of Bellman errors for value functions.
A complementary model-based perspective asks a different question: how efficiently can incorrect dynamics models be falsified?
If two candidate models make different predictions about transitions, there should be a ``witness'' test that distinguishes them, and the dimensionality of the space of such tests governs how quickly the algorithm can narrow down the correct model.

Sun et al.~\citep{sun2019witness} formalized this through witness rank, which measures the dimensionality of witnessed discrepancies between competing models.

\begin{definition}[Witness rank~\citep{sun2019witness}]
\label{def:witness-rank}
Let $\cM$ be a model class and $\cG$ a discriminator class.
The witness rank $W$ is the smallest integer such that, for each stage $h$, the discrepancy between any two models $M, M' \in \cM$ as measured by any test function $g \in \cG$ factors through $W$-dimensional embeddings of the state-action pair and the model pair.
\end{definition}

\begin{theorem}[Learning under witness rank]
\label{thm:witness}
With a realizable model class of witness rank $W$, model-based optimistic algorithms achieve $N(\varepsilon,\delta) = \tO(\mathrm{poly}(W, H) \cdot \varepsilon^{-2} \cdot \log(1/\delta))$ by iteratively ruling out models that are witnessed inconsistent with data~\citep{sun2019witness}.
\end{theorem}

Bellman rank and witness rank are incomparable in general: neither subsumes the other.
Both reduce to $S$ in the tabular case (Bellman rank with indicator features, witness rank when the model class contains all tabular MDPs), but they measure different aspects of the problem.
The choice between them reflects a modeling decision.
Value-based approaches (Bellman rank) support hypothesis elimination without an explicit dynamics model, making them natural when the value class is well-specified but the transition model is not.
Model-based approaches (witness rank) use moment-matching tests against a discriminator class $\cG$, and can be more statistically efficient when a good discriminator class is available and the model class is well-specified.
In practice, the distinction matters most when one representation (values vs.\ dynamics) is simpler than the other.

\subsection{The Bellman-Eluder dimension: an information-theoretic synthesis}

Jin, Liu, and Miryoosefi~\citep{jin2021bellmaneluder} unified and generalized the preceding notions through the Bellman-Eluder (BE) dimension, which measures the sequential complexity of a value class with respect to Bellman errors.
The definition adapts the eluder dimension from the bandit literature to the RL setting, capturing how quickly new observations resolve uncertainty about Bellman backups.

The core idea is independence in the Bellman-error sense.
A sequence of state-action queries is BE-independent if, at each point in the sequence, there exist two value functions in $\cF$ that agree on all previous queries (their cumulative Bellman errors are small) but disagree sharply at the current query.
The length of the longest such sequence is the BE dimension: it counts how many ``surprise'' observations the environment can produce before the algorithm must have identified the correct value function.

\begin{definition}[Bellman-Eluder dimension~\citep{jin2021bellmaneluder}]
\label{def:be-dim}
Given a value class $\cF$ and scale $\alpha > 0$, a sequence of queries $(x_1, a_1), \ldots, (x_n, a_n)$ is BE-independent if, for each $t$, there exist $f, f' \in \cF$ whose Bellman predictions at $(x_t, a_t)$ differ by at least $\alpha$ while their cumulative Bellman errors at all earlier queries remain small.
The BE dimension $d_{\mathrm{BE}}(\alpha)$ is the maximal length of such a sequence; write $d_{\mathrm{BE}} = \sup_{\alpha \in (0,1]} d_{\mathrm{BE}}(\alpha)$.
\end{definition}

\begin{theorem}[Learning with finite BE dimension]
\label{thm:be-structure}
If a value class $\cF$ has $d_{\mathrm{BE}} < \infty$, PAC and regret bounds scale polynomially in $(H, d_{\mathrm{BE}})$ with $\varepsilon^{-2}$ dependence~\citep{jin2021bellmaneluder}.
\end{theorem}

The BE dimension is the most general single-number complexity measure in the survey: it is finite whenever Bellman rank, witness rank, or bilinear rank is finite, and it equals $SA$ for tabular value classes.
The inclusion hierarchy is strict (there exist function classes with finite $d_{\mathrm{BE}}$ but infinite Bellman or witness rank), making it the broadest sufficient condition for PAC learnability with function approximation.
The tradeoff is precision: because $d_{\mathrm{BE}}$ accommodates such a wide range of function classes, the resulting bounds have less explicit dependence on problem parameters than Bellman-rank or linear-MDP bounds.

\subsection{Bilinear classes: a unifying factorization}

Du et al.~\citep{du2021bilinear} proposed bilinear classes as a structural framework that unifies several rank-based settings under a single algebraic condition.

\begin{definition}[Bilinear class~\citep{du2021bilinear}]
\label{def:bilinear}
A sequential decision model is bilinear of dimension $r$ if the Bellman backup residuals admit a rank-$r$ factorization through stage-dependent embeddings $\phi_h : \cX \times \cA \to \R^r$ and $\psi_h : \cX \to \R^r$ with bounded operator norms.
\end{definition}

The bilinear condition is a matrix-rank constraint on the interaction between state-action features and next-state features.
When this rank is small, the algorithm can estimate the Bellman backup using a low-dimensional inner product rather than a full nonparametric regression, and the sample complexity scales with $r$ rather than with the size of the observation space.

\begin{theorem}[Learning in bilinear classes]
\label{thm:bilinear}
Bilinear classes of dimension $r$ admit polynomial-time algorithms with $N(\varepsilon,\delta) = \tO(\mathrm{poly}(r, H) \cdot \varepsilon^{-2} \cdot \log(1/\delta))$, recovering and unifying guarantees for several previously studied feature-based and low-rank models~\citep{du2021bilinear}.
\end{theorem}

The bilinear framework recovers linear MDPs (where $\phi_h$ is the known feature map), low-rank MDPs (where $\phi_h$ and $\mu_h$ provide the factorization), and several other special cases as instances.
Its value is primarily conceptual: it shows that a single algebraic structure underlies many apparently different tractability results.

\subsection{The hierarchy and its practical meaning}

\begin{figure}[t]
\centering
\begin{tikzpicture}[scale=0.85, every node/.style={font=\small},
  box/.style={draw, rounded corners, minimum height=0.7cm, minimum width=1.8cm, align=center}]
  \node[box] (tab) at (0,0) {Tabular\\$SA$};
  \node[box] (lin) at (3,0) {Linear\\$d$};
  \node[box] (low) at (6,0) {Low-rank\\$r$};
  \node[box] (bil) at (9,0) {Bilinear\\$r$};
  \node[box] (be) at (12,0) {Finite $d_{\mathrm{BE}}$};
  \draw[-{Stealth},thick] (tab) -- (lin);
  \draw[-{Stealth},thick] (lin) -- (low);
  \draw[-{Stealth},thick] (low) -- (bil);
  \draw[-{Stealth},thick] (bil) -- (be);
  \node[below,font=\tiny,text=gray] at (1.5,-.55) {strict};
  \node[below,font=\tiny,text=gray] at (4.5,-.55) {strict};
  \node[below,font=\tiny,text=gray] at (7.5,-.55) {strict};
  \node[below,font=\tiny,text=gray] at (10.5,-.55) {strict};
  \node[above,font=\tiny] at (0,.55) {Tightest rates};
  \node[above,font=\tiny] at (12,.55) {Broadest scope};
\end{tikzpicture}
\caption{\textbf{Structural complexity hierarchy.} Each inclusion is strict: moving right trades tighter constants for broader applicability. The capacity parameter replacing $SA$ is shown below each class name.}
\label{fig:structure-hierarchy}
\end{figure}
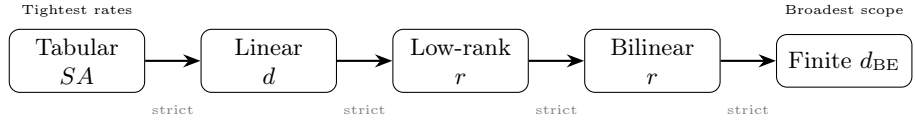

\begin{table}[t]
\centering
\small
\setlength{\tabcolsep}{4pt}
\begin{tabular}{lllll}
\toprule
\textbf{Measure} & \textbf{Type} & \textbf{Tabular value} & \textbf{Key assumption} & \textbf{Reference} \\
\midrule
Bellman rank $B$ & Value-based & $S$ & Realizability & \citet{jiang2017olive} \\
Witness rank $W$ & Model-based & $S$ & Realizable model class & \citet{sun2019witness} \\
Bilinear rank $r$ & Unified & $S$ & Rank-$r$ factorization & \citet{du2021bilinear} \\
BE dimension $d_{\mathrm{BE}}$ & Information-theoretic & $SA$ & Finite BE dimension & \citet{jin2021bellmaneluder} \\
\bottomrule
\end{tabular}
\caption{\textbf{Structural complexity measures compared.} All reduce to $S$ or $SA$ in the tabular case. The BE dimension is the most general: finite $B$, $W$, or bilinear rank each imply finite $d_{\mathrm{BE}}$, but not conversely.}
\label{tab:structure-compare}
\end{table}

Figure~\ref{fig:structure-hierarchy} and Table~\ref{tab:structure-compare} summarize the relationships.
The hierarchy of inclusions is
\[
\text{Tabular} \;\subset\; \text{Linear MDP} \;\subset\; \text{Low-rank MDP} \;\subset\; \text{Bilinear class} \;\subset\; \text{Finite } d_{\mathrm{BE}},
\]
where each inclusion is strict.
Moving right in this hierarchy trades specificity for generality: tabular bounds have the tightest constants and horizon exponents ($H^3$), while $d_{\mathrm{BE}}$-based bounds apply to the broadest class of problems but with less explicit parameter dependence.

For a practitioner confronting a new problem, the hierarchy suggests a diagnostic sequence.
Start by checking whether the problem is tabular (finite and small $S, A$); if so, use the minimax bounds of \S\ref{sec:tabular}.
If $S$ is large but features exist, test linear realizability using Bellman residual diagnostics (\S\ref{subsec:misspec}) and apply linear MDP bounds with $d \ll SA$.
If transitions factor through latent states, check low-rank or Block MDP structure (\S\ref{sec:rich}).
For general function classes where none of these special structures hold, compute or bound $d_{\mathrm{BE}}$ to obtain the most general guarantee.

The diagnostic sequence is monotone: each step relaxes assumptions and broadens applicability at the cost of weaker constants.
A practitioner should stop at the most specific level that fits, because tighter structural assumptions yield tighter rates.
If the linear residual test passes, there is no reason to invoke the more general $d_{\mathrm{BE}}$ framework; but if it fails, the broader framework provides a fallback.

\subsection{Computational considerations: what rates do not buy you}

Statistical learnability (small $B$, $W$, $d_{\mathrm{BE}}$, or $r$) does not automatically yield polynomial-time algorithms.
The gap between information-theoretic and computational feasibility is mediated by oracle efficiency: whether the algorithm's runtime is polynomial in the cost of standard supervised-learning subroutines (cost-sensitive classification or regression over the function class $\cF$).

Dann et al.~\citep{dann2018oracle} formalized this notion for rich-observation RL.
Bellman-rank elimination is typically oracle-efficient when the supervised subproblems are convex (e.g., ridge regression), but can become computationally hard when $\cF$ involves nonconvex constraints.
Witness-rank tests reduce to moment-matching over a discriminator class $\cG$, and the optimization difficulty scales with the richness of $\cG$.
In the agnostic setting, where realizability fails and the algorithm must compete with the best function in $\cF$, statistical-computational gaps are expected, and polynomial-time learning may require additional structural assumptions beyond finite rank or dimension.

The practical upshot is concrete.
When the function class is a set of linear models, the oracle is ridge regression: cheap, well-understood, and convex.
When the function class is a set of neural networks, the oracle is SGD on a nonconvex loss: practical in many cases but without worst-case guarantees.
When the function class is defined implicitly (e.g., all functions with bounded RKHS norm), the oracle may require kernel methods with $O(n^3)$ cost per call, which limits scalability.
A reader should treat the sample complexity bounds in this section as information-theoretic targets and assess computational feasibility separately for their specific function class and oracle.


\section{Function Approximation: From Linear Features to Neural Tangent Kernels}
\label{sec:function}

The structural measures of the previous section characterize when function approximation admits PAC learning; this section addresses how, under three increasingly general realizability assumptions.
We trace a progression from linear MDPs, where features are known and bounds are the sharpest, through reproducing kernel Hilbert spaces (RKHS), where the effective dimension replaces the feature count, to over-parameterized neural networks operated in the neural tangent kernel (NTK) regime, where kernel theory provides surrogate guarantees.

The common thread is that sample complexity replaces the tabular factor $SA$ with a dimension parameter ($d$ for linear features, $d_{\mathrm{eff}}(\lambda)$ for kernels) while preserving the $\varepsilon^{-2}$ core and introducing typically higher horizon exponents due to correlated estimation errors across the feature space.

\subsection{Linear MDPs: the cleanest structured setting}

The linear MDP framework, introduced by Jin, Yang, and Wang~\citep{jin2020linear}, posits that both rewards and transition expectations are linear in a known feature map.
This is the simplest nontrivial function approximation regime: the features are given, the regression is convex, and the resulting algorithms have explicit, polynomial-time implementations.

\begin{definition}[Linear MDP realizability~\citep{jin2020linear}]
\label{def:linear-mdp}
There exists a known feature map $\phi : \cS \times \cA \to \R^d$ with $\|\phi(s,a)\|_2 \le 1$ such that $r_h(s,a) = \phi(s,a)^\top \theta_h^{(r)}$ and $\E[g(s_{h+1}) \mid s,a] = \phi(s,a)^\top \theta_{h,g}$ for all bounded $g$, with $\|\theta\|_2 \le 1$.
\end{definition}

Under this assumption, the algorithm LSVI-UCB (Least-Squares Value Iteration with Upper Confidence Bounds) works as follows.
At each episode, it fits a ridge regression to estimate the $Q$-function at each stage, computes an elliptical confidence bonus that inflates the estimate in poorly explored directions of the feature space, and plays the greedy policy with respect to the optimistic estimate.
The elliptical bonus ensures that the algorithm explores directions in feature space where uncertainty is large, driving down estimation error uniformly across the feature space over successive episodes.

\begin{theorem}[LSVI-UCB for linear MDPs]
\label{thm:linmdp-lsvi}
Under linear MDP realizability with dimension $d$, LSVI-UCB achieves
\[
\mathrm{Regret}(T) = \tO\!\big(\sqrt{d^3 H^3 T}\big)
\quad\text{and}\quad
N(\varepsilon,\delta) = \tO\!\bigg(\frac{d^3 H^4}{\varepsilon^2}\bigg),
\]
with polynomial-time computation via ridge regression~\citep{jin2020linear}.
\end{theorem}

The $H^4$ exponent is higher than the tabular $H^3$, and the reason is instructive.
In the tabular setting, estimation errors at different state-action pairs are independent: learning about one state tells you nothing about another.
In the linear setting, estimation errors are correlated: a poorly estimated direction in feature space affects all state-action pairs whose features project onto that direction.
When these correlated errors feed into the Bellman backup at the next stage, they compound in ways that independent errors do not, adding an extra factor of $H$ to the sample complexity.
Variance-aware estimators can partially mitigate this effect, but closing the gap between $H^4$ and $H^3$ in the linear setting remains open.

He, Zhou, and Gu~\citep{he2021uniformpaclinear} extended these results to the uniform-PAC criterion:

\begin{theorem}[Uniform-PAC for linear MDPs]
\label{thm:linmdp-unifpac}
Under linear realizability and Bellman completeness, there exist uniform-PAC algorithms with budgets $N(\varepsilon,\delta) = \tO(\mathrm{poly}(d, H) \cdot \varepsilon^{-2})$, implying high-probability regret via Theorem~\ref{thm:uniformpac-regret}.
\end{theorem}

Bellman completeness (Definition~\ref{def:bellman-completeness}), the requirement that $\mathcal{T}_h \cF_{\mathrm{lin}} \subseteq \cF_{\mathrm{lin}}$, plays an important role here.
Without it, the Bellman backup of a linear function may not be linear, and the algorithm must project back into the linear class at each stage, introducing projection errors that accumulate over the horizon.
Whether completeness holds depends on the interaction between the feature map and the transition kernel: for some MDPs and features it holds exactly, for others it fails badly, and for most practical problems it is somewhere in between.
The residual diagnostic in \S\ref{subsec:misspec} provides a way to check empirically whether completeness holds well enough for the linear guarantees to be informative.

\subsection{Kernel and RKHS models: trading dimension for spectrum}

When the value function is smooth but not linear in a known low-dimensional feature map, reproducing kernel Hilbert space (RKHS) methods provide a natural generalization.
The idea is to work in a potentially infinite-dimensional feature space implicitly defined by a kernel function, and to control complexity through the norm of the value function in this space rather than through the number of features.

Yang et al.~\citep{yang2020kernelnn} analyzed kernel-based RL and introduced the effective dimension as the relevant capacity measure.

\begin{definition}[RKHS value class and effective dimension~\citep{yang2020kernelnn}]
\label{def:rkhs}
For a kernel $\kappa$ with RKHS $\cH_\kappa$, define the value class $\cF_\kappa = \{f : f_h \in \cH_\kappa,\, \|f_h\|_{\cH_\kappa} \le B\}$.
Given a distribution $\mu$ over $\cX$ with kernel integral operator $\Sigma = \E_{x \sim \mu}[\kappa(x, \cdot) \otimes \kappa(x, \cdot)]$, the effective dimension at regularization $\lambda > 0$ is
$d_{\mathrm{eff}}(\lambda) = \mathrm{Tr}(\Sigma(\Sigma + \lambda I)^{-1})$.
\end{definition}

The effective dimension is best understood through its spectral interpretation.
If the kernel integral operator has eigenvalues $\sigma_1 \ge \sigma_2 \ge \cdots$, then $d_{\mathrm{eff}}(\lambda) = \sum_j \sigma_j / (\sigma_j + \lambda)$.
Each eigenvalue contributes close to $1$ if $\sigma_j \gg \lambda$ (the corresponding direction is well-represented in the data) and close to $0$ if $\sigma_j \ll \lambda$ (the direction is effectively regularized away).
The effective dimension therefore counts the number of ``active'' directions in the RKHS at regularization level $\lambda$, interpolating between $d$ for finite-dimensional kernels and a potentially much smaller quantity for kernels with rapid spectral decay.
For Matérn kernels, $d_{\mathrm{eff}}(\lambda) = \cO(\lambda^{-d_x/\nu})$ where $d_x$ is the input dimension and $\nu$ the smoothness parameter; for Gaussian kernels, $d_{\mathrm{eff}}$ grows only polylogarithmically in $1/\lambda$.

\begin{theorem}[Kernel RL with effective dimension]
\label{thm:kernel}
Under realizability ($V^\star \in \cF_\kappa$), Bellman completeness, and a spectral condition controlling $d_{\mathrm{eff}}(\lambda)$, optimistic algorithms achieve
\[
N(\varepsilon,\delta) = \tO\!\big(\mathrm{poly}(H) \cdot d_{\mathrm{eff}}(\lambda) \cdot \varepsilon^{-2}\big)
\quad\text{and}\quad
\mathrm{Regret}(K) = \tO\!\big(\mathrm{poly}(H) \cdot \sqrt{d_{\mathrm{eff}}(\lambda) \cdot K}\big).
\]
The horizon exponents range from $H^4$ to $H^6$ depending on spectral assumptions; optimal exponents remain open~\citep{yang2020kernelnn}.
\end{theorem}

The wide range of horizon exponents ($H^4$ to $H^6$) reflects the fact that kernel RL is less mature than its linear counterpart.
The gap arises from different ways of handling the infinite-dimensional feature space: tighter spectral assumptions yield lower exponents but apply to fewer kernels, while looser assumptions accommodate more kernels at the cost of worse horizon dependence.
Resolving the optimal exponent is one of the open problems discussed in \S\ref{sec:open}.

In practice, selecting the regularization parameter $\lambda$ and the RKHS norm budget $B$ requires validation.
A useful surrogate is the ridge leverage score $\tau_i = x_i^\top (X^\top X + \lambda I)^{-1} x_i$, which measures how much influence each data point has on the kernel regression fit.
The sum of leverage scores equals $d_{\mathrm{eff}}(\lambda)$, so monitoring the distribution of leverage scores provides a data-driven estimate of the effective dimension and can guide both algorithmic tuning and coverage assessment (\S\ref{subsec:coverage}).

\subsection{Over-parameterized networks via the NTK regime}

A natural question is whether neural-network-based RL inherits the kernel guarantees.
The answer is conditionally yes, through the neural tangent kernel (NTK) connection.

When a neural network is sufficiently wide and trained with small-step gradient descent from random initialization, its predictions evolve approximately linearly in parameters.
In this regime, the network's behavior coincides with kernel regression under a specific kernel, the neural tangent kernel $\kappa_{\mathrm{NTK}}$, introduced by Jacot, Gabriel, and Hongler~\citep{jacot2018ntk} and analyzed in the RL context by Yang et al.~\citep{yang2020kernelnn}.
The consequence is that the kernel RL guarantees of Theorem~\ref{thm:kernel} transfer directly, with $d_{\mathrm{eff}}(\lambda)$ computed for $\kappa_{\mathrm{NTK}}$.

\begin{theorem}[NTK regime guarantees]
\label{thm:ntk}
For networks in the NTK regime, the guarantees of Theorem~\ref{thm:kernel} apply with $d_{\mathrm{eff}}(\lambda)$ computed for $\kappa_{\mathrm{NTK}}$, up to approximation and optimization errors that vanish with increasing width.
\end{theorem}

This transfer is valuable as a theoretical anchor but limited in practice.
The NTK regime requires width much larger than the training set size, small learning rates, and no aggressive optimization tricks (momentum, learning rate warmup, weight decay schedules).
Most practical deep RL systems violate all of these conditions.
With moderate width or aggressive optimization, the network leaves the NTK regime, its effective kernel changes during training, and the assumptions underlying Theorem~\ref{thm:kernel} (realizability, Bellman completeness, fixed effective dimension) may all fail simultaneously.

The honest assessment is that for networks operating far from the NTK limit, PAC guarantees are largely absent.
The theory tells us what would need to be true (small $d_{\mathrm{eff}}$, approximate completeness, controlled Bellman errors) but does not certify that these conditions hold in a given deep RL pipeline.
Practitioners working with standard-width networks should treat the model as potentially misspecified and rely on alternative safeguards: pessimism in offline settings, off-policy evaluation for deployment decisions, PAC-Bayes bounds (\S\ref{sec:pacbayes}) for distributional guarantees, and the residual diagnostics of \S\ref{subsec:misspec} to flag when guarantees are likely to fail.

\subsection{Reward-free exploration with function approximation}

Reward-free exploration (\S\ref{sec:rfe}) extends naturally to the linear setting.
Wang et al.~\citep{wang2020linearRFE} showed that two-stage exploration under linear realizability achieves $\tO(\mathrm{poly}(d, H) \cdot \varepsilon^{-2})$ episodes to guarantee $\varepsilon$-optimality for any downstream reward, and Kaufmann and Tirinzoni~\citep{kaufmann2021adaptive} developed adaptive strategies that improve on worst-case budgets for benign instances.
Through the CSO lens, reward-free exploration creates a reusable coverage resource: the exploration budget (first stage) is an investment in $\mathsf{Cov}$ that amortizes across arbitrarily many rewards.

The kernel extension is more delicate.
Because $d_{\mathrm{eff}}(\lambda)$ depends on the data distribution (through the kernel integral operator), the effective dimension during exploration may differ from the effective dimension during planning, and the two-stage guarantee requires that the exploration policy creates a distribution under which $d_{\mathrm{eff}}$ is well-controlled.
This interplay between exploration design and spectral properties is not fully resolved and connects to the open problems on kernel uniform-PAC in \S\ref{sec:open}.

\subsection{From function approximation to offline RL}

Linear realizability also underpins the pessimistic offline RL guarantees of \S\ref{sec:offline}.
The structural parameter $d$ enters the offline sample complexity alongside the coverage coefficient $C_\star$, and the interplay between feature quality and data support determines whether offline policy improvement is feasible.

To see why both factors matter, consider two failure modes.
In the first, the features are correct ($Q^\star$ is linear in $\phi$) but the data has poor coverage ($C_\star$ is large): the algorithm cannot estimate $Q^\star$ accurately in the directions that matter for $\pi^\star$, and pessimism produces an overly conservative policy.
In the second, the data has excellent coverage ($C_\star$ is small) but the features are wrong ($Q^\star$ is not linear in $\phi$): the algorithm fits the wrong function confidently, and pessimism cannot protect against systematic bias in the wrong direction.
Only when both features and coverage are adequate can pessimistic offline RL deliver on its guarantees.
When features are misspecified, offline guarantees degrade in ways that additional data cannot repair, a theme we return to in the open problems of \S\ref{sec:open}.

\subsection{Practical guidance for function approximation}

The theory in this section generates several principles that inform applied work.

The most important principle is to verify assumptions before invoking bounds.
Linear MDP guarantees assume that rewards and transitions are linear in the provided features and that the function class is Bellman-complete.
If these conditions fail, the bounds are not merely loose; they can be actively misleading, giving false confidence in a policy that performs poorly.
The Bellman residual test in \S\ref{subsec:misspec} (Algorithm~\ref{alg:bellman}) is the first diagnostic to run: fit value functions via ridge regression on random-policy data, compute held-out Bellman residuals at each stage, and check whether residuals remain small and stable.
Growing or systematically biased residuals indicate misspecification.

For kernel pipelines, the choice of kernel and regularization parameter $\lambda$ determines the effective dimension and therefore the sample complexity.
Selecting $\lambda$ via cross-validation and tracking the distribution of ridge leverage scores provides a data-driven estimate of $d_{\mathrm{eff}}$ that can guide both algorithm design and coverage assessment.
If leverage scores are concentrated near $1$ for many data points, the effective dimension is large and more data may be needed; if they decay rapidly, the kernel is doing its job of compressing the problem.

In all function approximation settings, maintaining a well-conditioned design matrix is important for stable estimation.
The elliptical bonus in LSVI-UCB is not just a theoretical device but a practical necessity: it ensures that the algorithm does not repeatedly visit the same directions in feature space while ignoring others.
In offline settings where the agent cannot control exploration, checking the condition number of the design matrix at each stage provides a proxy for how well the data covers the feature space.

For long-horizon problems, monitor the growth of confidence radii with $H$.
If bounds inflate rapidly with the horizon, this signals that estimation errors are compounding through Bellman backups.
Remedies include state abstraction (reducing the effective state space), temporal hierarchy (breaking the problem into shorter sub-episodes), or increased discounting (reducing the effective horizon $H_{\mathrm{eff}}$).

Finally, for deep RL pipelines that do not fit cleanly into the linear or NTK regime, the most honest posture is epistemic caution.
Treat the network as a potentially misspecified function class, prefer pessimistic objectives in offline settings, use off-policy evaluation to check policy quality before deployment, and consider PAC-Bayes bounds (\S\ref{sec:pacbayes}) as a way to obtain distributional guarantees that do not require realizability.
The guarantees of this section provide targets and diagnostics, not blanket assurances.

\section{Rich Observations and Low-Rank Structure}
\label{sec:rich}

Many practical RL problems present the agent with high-dimensional observations that are far richer than the underlying dynamics require.
A robot arm receives camera images with millions of pixels, but its task-relevant state (joint angles, gripper position, object location) lives in a space of perhaps twenty continuous variables.
A patient in an ICU generates streams of lab panels, vital signs, and clinical notes, but the treatment-relevant state may reduce to a handful of latent physiological regimes (stable, deteriorating, septic, recovering).
In both cases, the observation space $\cX$ is enormous, but the transition dynamics depend on a compact latent representation, and the gap between observation complexity and dynamical complexity is what makes structured learning possible.

This section covers two formalizations of this idea: Block MDPs, where each observation reveals a discrete latent state through a decodable mapping, and low-rank MDPs, where the transition operator factors through a low-dimensional feature space without requiring discrete latent states.
Both reduce the effective state count from $|\cX|$ to a much smaller quantity ($m$ or $r$), and both connect to the structural complexity hierarchy of \S\ref{sec:structure}.

\subsection{Block MDPs: decodable latent states}

The Block MDP framework posits that observations are generated by a small number of latent states, and that an agent can learn to decode observations back to latent states from interaction data alone.

The setup is as follows.
The environment has a latent state space $[m] = \{1, \ldots, m\}$ with $m \ll |\cX|$.
At each stage, the latent state $z_h \in [m]$ determines both the transition to $z_{h+1}$ and the distribution over observations $x_h$.
The agent sees $x_h$ but not $z_h$; however, the observation distribution varies enough across latent states that a decoder $\hat\psi : \cX \to [m]$ can be learned with controlled error.

\begin{definition}[Block MDP]
\label{def:block-mdp}
A Block MDP has a surjective map $\psi : \cX \to [m]$ such that the transition kernel $P(x' \mid x, a)$ depends on $x$ only through $\psi(x)$, and the reward depends on $(\psi(x), a)$.
The observation distributions conditioned on different latent states are separable: for $z \neq z'$, the supports of $P(x \mid z)$ and $P(x \mid z')$ are disjoint (or, in relaxed versions, statistically distinguishable).
\end{definition}

Du et al.~\citep{du2019provable} initiated provably efficient RL with rich observations by showing that, under the decodability condition, the agent can learn a decoder from data and then plan in the latent MDP with $m$ states rather than in the full observation space.
Misra et al.~\citep{misra2020homer} developed HOMER, a representation-learning algorithm that combines contrastive learning (to discover the decoder) with optimistic exploration (to ensure coverage of all latent states), achieving sample complexity polynomial in $(m, A, H, 1/\varepsilon)$.

The key insight is that once a good decoder is available, the Block MDP reduces to a tabular MDP with $m$ states, and the PAC guarantees of \S\ref{sec:tabular} apply with $m$ replacing $S$.
The cost of learning the decoder is folded into the polynomial dependence on $m$: representation learning adds a multiplicative factor but does not change the fundamental scaling.

\subsection{Low-rank MDPs: continuous latent structure}

Block MDPs require discrete latent states with hard boundaries between regimes.
Low-rank MDPs relax this to a continuous factorization of the transition operator, accommodating settings where the latent structure is smooth rather than discrete.

\begin{definition}[Low-rank MDP]
\label{def:lowrank-mdp}
A low-rank MDP has, for each stage $h$, a factorization $P_h(s' \mid s, a) = \langle \phi_h(s, a),\, \mu_h(s') \rangle$ for bounded feature maps $\phi_h : \cS \times \cA \to \R^r$ and $\mu_h : \cS \to \R^r$, where $r$ is the rank.
\end{definition}

The factorization means that the transition operator at each stage has rank $r$ when viewed as a matrix (or operator) mapping state-action pairs to next-state distributions.
The features $\phi_h$ and $\mu_h$ may be unknown and must be learned from data, unlike the linear MDP setting of \S\ref{sec:function} where $\phi$ is given.

Agarwal et al.~\citep{agarwal2020flambe} analyzed model-based exploration with learned low-rank features through the FLAMBE algorithm, showing that representation learning and optimistic planning can be combined to achieve sample complexity polynomial in $(r, H, 1/\varepsilon)$.
Dann et al.~\citep{dann2021agnostic} studied the agnostic case, where the rank-$r$ assumption holds only approximately, and showed that existing algorithms incur an additive misspecification penalty that does not vanish with more data.
Huang et al.~\citep{huang2023density} proposed density-feature methods that exploit the factorization from the next-state side ($\mu_h$), providing an alternative to the state-action feature approach.

\begin{theorem}[Block and low-rank sample complexity]
\label{thm:block-lowrank}
Under realizability and identifiability (decodability for Block MDPs, or rank-$r$ factorization for low-rank MDPs), polynomial-time algorithms achieve
\[
N(\varepsilon,\delta) = \tO\!\big(\mathrm{poly}(H) \cdot \mathrm{poly}(m) \cdot A\, \varepsilon^{-2}\big)
\quad\text{(Block MDPs)}
\]
\[
\text{or}\quad
\tO\!\big(\mathrm{poly}(H) \cdot \mathrm{poly}(r)\, \varepsilon^{-2}\big)
\quad\text{(low-rank MDPs)},
\]
using representation learning and optimistic planning~\citep{du2019provable,misra2020homer,agarwal2020flambe,dann2021agnostic,huang2023density}.
\end{theorem}

\subsection{Connections to structural complexity and function approximation}

Both Block and low-rank MDPs connect to the structural complexity hierarchy of \S\ref{sec:structure} and the function approximation results of \S\ref{sec:function}.

When a low-dimensional latent structure exists, the Bellman rank and witness rank are typically small.
A Block MDP with $m$ latent states has Bellman rank at most $m$ (using latent-state indicators as features), and a low-rank MDP with rank $r$ has bilinear rank at most $r$.
This explains why the sample complexity scales with $m$ or $r$ rather than $|\cX|$: the structural complexity measures detect the latent simplicity even when the observation space is vast.

The relationship to linear and kernel function approximation is complementary rather than hierarchical.
Linear MDP features (Definition~\ref{def:linear-mdp}) are given and fixed; low-rank features must be learned.
When the low-rank features happen to be linear in a known basis, the low-rank setting reduces to the linear MDP setting and the guarantees of \S\ref{sec:function} apply directly.
When the features are nonlinear or unknown, representation learning adds a cost that the linear theory does not account for, but the resulting guarantees still scale with the intrinsic dimension $r$ rather than the ambient dimension $|\cX|$.

Through the CSO lens, both Block and low-rank MDPs have $\mathsf{Comp} = \mathrm{poly}(m)$ or $\mathrm{poly}(r)$, with $\mathsf{Cov} = 1$ under online access.
The partial observability (or high dimensionality) of the observation space is absorbed entirely into the structural term: it affects which algorithms are needed (those with representation learning) but not the fundamental scaling of sample complexity.

\subsection{Practical diagnostics for latent structure}
\label{subsec:latent-diagnostics}

An applied researcher facing a high-dimensional observation space should look for evidence of latent structure before committing to a Block or low-rank model.

The most accessible diagnostic is the empirical rank of the transition operator.
Collect a dataset of transitions $(x_h, a_h, x_{h+1})$, compute a kernel or neural embedding of each tuple, and examine the singular value spectrum of the resulting matrix.
A rapid spectral decay, with most variance captured by a few singular values, suggests low-rank structure; a flat spectrum suggests that the observations carry genuinely high-dimensional information and that Block or low-rank models may be inappropriate.

If spectral evidence is encouraging, test whether simple function approximation suffices in the compressed space.
Fit a linear $Q$-function head on top of a learned embedding and evaluate prediction accuracy on held-out data.
If the linear head predicts returns well, the effective latent dimension is low and the guarantees of this section or of \S\ref{sec:function} are likely to apply.
If the linear head fails but a nonlinear head succeeds, latent structure exists but may require kernel or neural representation learning to exploit.

For settings where Block MDP structure is specifically hypothesized, contrastive state decoding provides a targeted test.
Train a classifier to predict whether two observations $x, x'$ come from the same latent state, using temporal adjacency or reward similarity as supervision signals.
High classifier accuracy with a simple architecture confirms that the Block structure is present and decodable; poor accuracy suggests that latent states may not be identifiable from observations alone, and the practitioner should consider the more general low-rank framework or fall back to $d_{\mathrm{BE}}$-based guarantees from \S\ref{sec:structure}.

When diagnostics suggest that latent structure is present, prefer algorithms that learn a decoder or fit a factored model before performing value iteration, and use per-episode policy certificates (Definition~\ref{def:cert}) to gate deployment.
When diagnostics are ambiguous, the Bellman-Eluder dimension framework of \S\ref{sec:structure} provides guarantees without committing to a specific latent model, though at the cost of less explicit parameter dependence.

\section{Reward-Free Exploration}
\label{sec:rfe}

A recurring theme in the preceding sections is that sample complexity depends on alignment between the data-collection process and the target policy.
In standard online RL, the agent collects data while pursuing a single reward signal, and exploration is tailored to that reward.
But what happens when the reward is not available during data collection, or when the same dataset must support planning for many different rewards?

This question arises naturally in several practical settings.
In multi-task robotics, an engineer may want to train a single robot arm to perform dozens of manipulation tasks, each defined by a different reward function, without re-exploring the environment for each task.
In safety auditing, a regulator may need to evaluate a deployed system's behavior under multiple safety criteria after data has already been collected.
In drug development, clinical trial data gathered under one primary endpoint may need to be reanalyzed for secondary endpoints that were not specified at design time.
In all these cases, the cost of data collection is high and the reward of interest is either unknown or plural at collection time.

Reward-free exploration (RFE) formalizes this problem.
Jin et al.~\citep{jin2020rewardfree} introduced a two-stage protocol that separates data collection from reward exploitation, and established the first near-optimal sample complexity bounds for the tabular case.

\subsection{The two-stage protocol}

RFE algorithms operate in two phases.
In the exploration phase, the agent interacts with the environment without access to any reward signal.
Its goal is to collect enough data (or build a sufficiently accurate model) that the resulting information supports near-optimal planning for any reward function that might be specified later.
In the planning phase, a reward function $r \in [0,1]$ is revealed, and the agent must output an $\varepsilon$-optimal policy for that reward using only the data or model from the first phase, with no further interaction.

The exploration phase faces a fundamentally harder task than standard online exploration.
In standard RL, the agent only needs to visit states and actions that are relevant to the specific reward it is optimizing.
In RFE, the agent must visit states and actions that might be relevant to any reward, because the reward is not yet known.
This means the exploration phase must achieve broad coverage of the state-action occupancy measures: it must ensure that every state-action pair reachable under some policy is visited often enough to estimate its transition dynamics accurately.

The cost of this broadened requirement appears as an extra factor in the sample complexity.
In the tabular setting, standard online PAC learning requires $\tTheta(SAH^3/\varepsilon^2)$ episodes for a single reward.
Reward-free exploration requires $\tO(S^2 A\,\mathrm{poly}(H)/\varepsilon^2)$, with an extra factor of $S$.
The additional $S$ arises because the agent must cover the occupancy measures for all $S$ possible ``target states'' that a future reward might make important, not just the states relevant to one particular reward.
This is a coverage investment: the exploration phase pays $S$ upfront so that the planning phase can handle any reward for free.

\subsection{Tabular guarantees}

\begin{theorem}[Tabular RFE]
\label{thm:rfe-tab-dep}
There exist algorithms that, without accessing rewards, collect $\tO(S^2 A\,\mathrm{poly}(H)\,\varepsilon^{-2})$ episodes and subsequently, for any reward $r \in [0,1]$, compute an $\varepsilon$-optimal policy with probability $1 - \delta$~\citep{jin2020rewardfree}.
\end{theorem}

The matching lower bound of $\Omega(S^2 A H^2 / \varepsilon^2)$ confirms that the $S^2$ factor is information-theoretically necessary in the reward-free setting~\citep{jin2020rewardfree}.
No algorithm can avoid this cost: if the exploration phase does not visit some state $s$ often enough, there exists a reward function that places all its weight on $s$, making the planning phase fail.

Through the CSO lens, the tabular RFE bound parses as $\mathsf{Cov} = S$ (the coverage investment for supporting all rewards), $\mathsf{Comp} = SA$ (tabular structure), and the standard $\varepsilon^{-2}$ core.
Compared to single-reward online learning, the only change is the coverage axis: the extra factor of $S$ is the price of reward generality.

It is worth asking when this price is worth paying.
If the agent will face only one reward, standard online learning is cheaper by a factor of $S$.
But if the agent will face $k$ different rewards sequentially, standard online learning costs $k \cdot \tTheta(SAH^3/\varepsilon^2)$ while reward-free exploration costs $\tO(S^2 A\,\mathrm{poly}(H)/\varepsilon^2)$ regardless of $k$.
The crossover point is roughly $k \approx S$: once the number of downstream tasks exceeds the number of states, the upfront RFE investment is amortized and becomes cheaper than repeated single-task exploration.

\subsection{Extension to linear function approximation}

The two-stage protocol extends naturally to settings with function approximation, where the feature dimension $d$ replaces the state count $S$ in the complexity bound.

\begin{theorem}[Linear RFE]
\label{thm:rfe-lin}
Under linear MDP realizability with feature dimension $d$, two-stage RFE achieves $\tO(\mathrm{poly}(H)\,\mathrm{poly}(d)\,\varepsilon^{-2})$ samples to guarantee $\varepsilon$-optimality for any reward, with probability $1 - \delta$~\citep{wang2020linearRFE,kaufmann2021adaptive}.
\end{theorem}

Wang et al.~\citep{wang2020linearRFE} established this result by showing that the exploration phase can build an accurate linear model of the transition dynamics (using elliptical exploration bonuses similar to those in LSVI-UCB) and that the planning phase can extract near-optimal policies for any reward by solving a linear program over the estimated model.
The sample complexity depends polynomially on $d$ and $H$ but not on $S$, which can be infinite in the linear setting.

Kaufmann and Tirinzoni~\citep{kaufmann2021adaptive} improved on the worst-case budget with adaptive strategies that allocate exploration effort based on the observed difficulty of different regions of the feature space.
Their algorithm collects fewer samples in well-understood directions and concentrates effort on directions where the model remains uncertain, achieving instance-dependent rates that can be substantially better than the worst-case bound for benign problems.

The kernel extension of RFE is more subtle.
Because the effective dimension $d_{\mathrm{eff}}(\lambda)$ depends on the data distribution (through the kernel integral operator), the effective dimension during exploration may differ from the effective dimension relevant to planning under a specific reward.
The exploration phase must create a distribution under which $d_{\mathrm{eff}}$ is well-controlled for all possible planning distributions, and this requires careful design of the exploration policy.
This interplay between exploration design and spectral properties is not fully resolved and connects to the open problems on kernel uniform-PAC in \S\ref{sec:open}.

\subsection{Connections to other settings}

Reward-free exploration sits at the intersection of online learning and offline RL in an instructive way.

In standard online RL, the agent constructs coverage implicitly: exploration bonuses steer it toward informative states, and coverage builds up as a byproduct of reward optimization.
In offline RL, coverage is fixed by the behavior policy and the agent must cope with whatever support the data provides.
RFE occupies the middle ground: the agent has online access (it can explore freely) but must construct coverage explicitly and deliberately, because the reward that will determine what counts as ``informative'' is not yet known.

This positioning makes RFE a natural preprocessing step for offline RL pipelines.
If an organization can collect data through active exploration before committing to a specific reward function, the RFE protocol ensures that the resulting dataset has the coverage properties needed for downstream offline learning.
The exploration phase creates broad coverage, and the planning phase (or a subsequent offline RL algorithm) exploits it.
In the CSO framework, RFE transforms a problem with $\mathsf{Cov} = \mathrm{poly}(C_\star)$ (uncertain offline coverage) into one with $\mathsf{Cov} = 1$ (guaranteed coverage), at the cost of the exploration budget.

\subsection{Practical considerations}

Several practical considerations guide the use of RFE in applied settings.

The most important question is whether the number of downstream tasks justifies the upfront exploration cost.
If the agent will optimize a single known reward, standard online learning is more efficient.
If the agent will face many rewards, or if the reward will be revised during deployment (as in iterative product design or adaptive clinical trial protocols), RFE amortizes the exploration cost and avoids repeated data collection.

During the exploration phase, maintaining well-conditioned design matrices is essential for efficient coverage creation.
In the linear setting, the elliptical exploration bonus (the same mechanism used in LSVI-UCB) ensures that the algorithm does not over-explore familiar directions while ignoring uncertain ones.
In practice, monitoring the eigenvalues of the design matrix $X_h^\top X_h + \lambda I$ at each stage provides a diagnostic: if the smallest eigenvalue remains close to $\lambda$ (the regularization floor) after many episodes, some directions in feature space are still unexplored, and the exploration phase should continue.

For long-horizon problems, the exploration budget grows with $\mathrm{poly}(H)$, which can be prohibitive.
Strategies for managing this cost include temporal abstraction (breaking the problem into shorter sub-episodes and exploring each independently), state abstraction (reducing the effective state space before exploration), and staged exploration (exploring coarsely first, then refining coverage in regions where early planning reveals gaps).

When the exploration phase ends and a reward is revealed, the quality of the resulting policy depends on how well the exploration data covers the states and actions relevant to that specific reward.
If the reward happens to concentrate on a small region of the state space that was well-explored, the planning phase will produce a near-optimal policy with minimal additional computation.
If the reward depends on a region that was poorly covered despite the algorithm's best efforts (perhaps because it is hard to reach), the policy may be suboptimal, though still within the $\varepsilon$ guarantee provided by the theorem.
Monitoring the estimated model uncertainty in reward-relevant regions provides a diagnostic for this: low uncertainty indicates that the exploration budget was well-spent for this particular reward, while high uncertainty suggests that a larger exploration budget or a more targeted exploration strategy would have been beneficial.

\section{Offline RL: Learning from Fixed Data}
\label{sec:offline}

In all the settings considered so far, the agent can collect new data by interacting with the environment.
Online exploration and reward-free exploration both assume that the agent controls its own data-generating process, steering it toward informative regions of the state-action space through optimistic bonuses or deliberate coverage-building.
Offline RL removes this assumption entirely.
The agent receives a fixed dataset $\cD$ of trajectories collected by some behavior policy $\mu$, and must learn the best policy it can without any further interaction with the environment.

This setting is relevant whenever additional data collection is impossible, expensive, or dangerous.
A hospital cannot run new experiments on past patients to improve a treatment policy; it must learn from the medical records already on file.
An autonomous driving company cannot redeploy a potentially unsafe controller to collect better training data; it must extract a policy from logs of human driving.
A recommendation system cannot repeatedly show bad recommendations to users to explore the reward landscape; it must improve using historical click data.

The central challenge of offline RL is coverage.
The dataset $\cD$ reflects the behavior policy $\mu$, which may visit very different states and actions than the optimal policy $\pi^\star$.
If $\mu$ never visits a state that $\pi^\star$ would, the dataset contains no information about what to do there, and no algorithm can reliably improve on $\mu$ in that region.
The concentrability coefficient $C_\star$ (Definition~\ref{def:concentrability}) quantifies this mismatch: when $C_\star$ is small, the data adequately covers the optimal policy's trajectory; when $C_\star$ is large, critical regions are under-represented; when $C_\star$ is infinite, improvement is impossible without further assumptions.

\subsection{Pessimism as a design principle}

In online RL, the standard strategy is optimism: inflate value estimates in uncertain regions to encourage the agent to visit them and collect informative data.
In offline RL, the agent cannot visit uncertain regions because it cannot collect new data.
Optimism in this setting is counterproductive: it overestimates the value of actions the data says little about, leading to policies that confidently choose poorly supported actions and fail on deployment.

Pessimism inverts this logic.
Instead of inflating estimates in uncertain regions, pessimistic algorithms deflate them, subtracting a penalty proportional to the estimation uncertainty at each state-action pair.
The effect is that the algorithm trusts the data where support is strong and defaults to conservative (low-value) estimates where support is weak.
The resulting policy avoids regions where the data is uninformative, even if the true value in those regions might be high.

More concretely, pessimistic value iteration (PEVI), introduced by Jin, Yang, and Wang~\citep{jin2021pevi}, modifies the standard Bellman backup by subtracting a confidence bonus rather than adding one:
\[
\hat{Q}_h(s,a) = \hat{r}_h(s,a) + \hat{P}_h \hat{V}_{h+1}(s,a) - \Gamma_h(s,a),
\]
where $\hat{r}_h$ and $\hat{P}_h$ are estimates of the reward and transition from data, and $\Gamma_h(s,a)$ is a penalty that grows with the uncertainty of the estimate at $(s,a)$.
In well-covered regions (where $\mu$ visits $(s,a)$ frequently), $\Gamma_h$ is small and the estimate is close to the true value.
In poorly covered regions (where $\mu$ rarely visits $(s,a)$), $\Gamma_h$ is large and the estimate is pushed down, discouraging the policy from relying on those actions.

Pessimistic Q-learning (PQL), analyzed by Shi et al.~\citep{shi2022pql}, applies the same principle in a model-free setting, subtracting penalties directly from Q-value estimates without building an explicit transition model.

\subsection{Guarantees with linear structure}

The formal guarantees for pessimistic offline RL combine the structural parameter $d$ (from the function class) with the coverage coefficient $C_\star$ (from the data distribution).

\begin{theorem}[Linear MDPs: pessimistic control]
\label{thm:offline-lin}
Under linear MDP realizability with feature dimension $d$ and concentrability $C_\star$, pessimistic algorithms (PEVI/PQL) achieve control error $\varepsilon$ with dataset size
\[
n \;=\; \tO\!\Big(\mathrm{poly}(H)\, \mathrm{poly}(d)\, \mathrm{poly}(C_\star)\, \varepsilon^{-2}\Big),
\]
with polynomial-time computation~\citep{jin2021pevi,shi2022pql}.
\end{theorem}

The bound reveals an important interaction between the CSO axes.
The structural term $\mathrm{poly}(d)$ reflects the complexity of the function class, just as in the online setting.
The coverage term $\mathrm{poly}(C_\star)$ is new: it measures how much the data distribution deviates from what the optimal policy needs, and it can dominate the bound entirely.
Even with $d = 5$ (a simple linear model), if the behavior policy barely overlaps with $\pi^\star$ (say, $C_\star = 10^4$), the required dataset size becomes enormous.
Conversely, if the behavior policy happens to be close to optimal ($C_\star$ near $1$), offline learning is almost as efficient as online learning.

The practical lesson is that coverage, not model complexity, is usually the binding constraint in offline RL.
A practitioner with a rich dataset that was collected by a diverse behavior policy (small $C_\star$) is in a strong position regardless of the model class.
A practitioner with a narrow dataset (large $C_\star$) will struggle no matter how sophisticated the algorithm.

\subsection{Model-based offline RL in the tabular setting}

When the state and action spaces are finite, model-based methods provide an alternative to value-based pessimism.
The idea is to estimate the transition model $\hat{P}_h$ and reward model $\hat{r}_h$ from data, quantify the uncertainty in these estimates, and plan pessimistically in the estimated model by penalizing transitions through poorly estimated regions.

\begin{theorem}[Tabular offline sample complexity]
\label{thm:offline-tab}
There exist model-based offline procedures whose sample complexity to achieve $\varepsilon$-optimal control is minimax-optimal (up to logarithms) with explicit dependence on $(S, A, H)$ and coverage~\citep{li2024aos}.
\end{theorem}

Li et al.~\citep{li2024aos} settled the minimax sample complexity in this setting, showing that model-based pessimism is not only sound but optimal: no algorithm can do better in the worst case.
The result confirms that the concentrability coefficient is the right measure of coverage difficulty in the tabular offline regime, and that the multiplicative structure of the CSO template (coverage times structure times accuracy) is tight.

\subsection{Off-policy evaluation}

Before deploying an offline-learned policy, a practitioner may want to estimate its value without actually running it in the environment.
Off-policy evaluation (OPE) addresses this: given a dataset collected under $\mu$ and a target policy $\pi$, estimate $V_1^\pi(\rho)$.

OPE is a strictly easier task than policy optimization.
Evaluation asks for an accurate value estimate; optimization asks for a policy that achieves high value.
Evaluation can succeed even when optimization cannot, because estimating the value of a specific policy requires coverage only along that policy's trajectory, not along the optimal policy's trajectory.
This distinction matters in practice: when coverage is too poor for safe policy improvement, OPE may still provide useful information about candidate policies.

Two families of estimators are widely used.
Importance-weighted estimators re-weight the observed returns by the ratio $\pi / \mu$ of the target and behavior policy probabilities.
Thomas, Theocharous, and Ghavamzadeh~\citep{thomas2015hcope,thomas2016doublyrobust} developed high-confidence OPE using importance weighting with doubly-robust corrections that reduce variance.
Doubly-robust and semiparametric efficient estimators combine a model-based estimate with importance weighting to achieve the lowest possible asymptotic variance under standard regularity conditions.
Kallus and Uehara~\citep{kallus2020ope} established the efficiency theory for this approach, and Jiang and Huang~\citep{jiang2020minimaxinterval} showed how to construct minimax value intervals with finite-sample validity.

\begin{theorem}[Efficient OPE]
\label{thm:ope}
Under mild regularity, there exist OPE estimators that achieve statistical efficiency, and minimax value intervals can be constructed with finite-sample validity~\citep{kallus2020ope,jiang2020minimaxinterval}.
\end{theorem}

The practical value of OPE extends beyond point estimation.
Interval estimates (confidence intervals for $V_1^\pi(\rho)$) tell the practitioner not just how good the policy might be, but how uncertain the estimate is.
A narrow interval centered above the behavior policy's value suggests safe deployment; a wide interval or one overlapping zero suggests that the data does not support confident claims about the policy's quality.
This makes OPE a natural deployment gate, complementing the coverage diagnostics of \S\ref{subsec:coverage} and the policy certificates of Definition~\ref{def:cert}.

\subsection{The CSO view of offline RL}

Through the CSO lens, offline RL is the setting where the coverage axis dominates.

In the linear case, $\mathsf{Comp} = \mathrm{poly}(d)$ is the same structural term that appears in online learning, and $\mathsf{Obj}$ is pessimistic control or evaluation.
The new element is $\mathsf{Cov} = \mathrm{poly}(C_\star)$, which can be the largest factor in the bound.
In the tabular model-based case, $\mathsf{Comp} = SA$ and $\mathsf{Cov} = \mathrm{poly}(C_\star)$, and the bound is minimax-optimal.

This parsing suggests a diagnostic workflow for offline RL.
First, estimate a coverage proxy $\widehat{C}_\star$ using the procedures in \S\ref{subsec:coverage} (density ratios, ridge leverage scores, effective sample sizes).
If the proxy is small (say, $\widehat{C}_\star < 30$), the data has reasonable coverage and pessimistic control algorithms like PEVI or PQL are appropriate.
If the proxy is moderate ($30 < \widehat{C}_\star < 100$), proceed with caution: use pessimism but validate the resulting policy with OPE before deployment.
If the proxy is large ($\widehat{C}_\star > 100$) or the density ratios have heavy tails, the data does not adequately support policy improvement.
In this regime, prefer OPE interval estimates over policy optimization, and abstain from deployment until additional data can be collected or the target policy can be shrunk toward the behavior policy (reducing $C_\star$ at the cost of optimality).

\subsection{Two failure modes and when to worry}

Two distinct failure modes affect offline RL, and distinguishing between them is essential for diagnosis.

The first is coverage failure: the function class is correct ($Q^\star \in \cF$) but the data distribution is poorly aligned with $\pi^\star$.
In this case, the algorithm cannot estimate $Q^\star$ accurately in the directions that matter, and pessimism produces an overly conservative policy that avoids high-value but under-supported actions.
The remedy is more or better data: either collect additional trajectories that cover the missing regions, or modify the behavior policy to be more exploratory.

The second is misspecification failure: the data has adequate coverage ($C_\star$ is small) but the function class is wrong ($Q^\star \notin \cF$).
In this case, the algorithm fits the wrong function confidently, and pessimism cannot protect against systematic bias because the bias does not manifest as uncertainty in the covered directions.
The remedy is a richer function class or a model-based approach that is less sensitive to value-function misspecification.

When both failure modes are present simultaneously (poor coverage and misspecified features), the situation is particularly difficult.
The total error decomposes into approximation error (from misspecification), estimation error (from finite data), and coverage error (from distribution mismatch), but these three components interact in ways that current theory does not fully characterize.
Obtaining sharp bounds on this three-way tradeoff with tractable algorithms is one of the frontier open problems identified in \S\ref{sec:open}.

\subsection{Practical guidance for offline pipelines}

The theory in this section leads to a concrete workflow for offline RL deployment.

Before running any offline algorithm, estimate coverage using the diagnostics in \S\ref{subsec:coverage}.
Compute density ratios between a candidate policy and the behavior policy at each stage, clipping extreme values to stabilize the estimates.
Report effective sample sizes and upper quantiles of the importance weights.
If coverage is adequate, proceed to pessimistic learning; if not, assess whether OPE alone can answer the relevant question without policy improvement.

After running a pessimistic algorithm, validate the learned policy with OPE before deployment.
Compute a confidence interval for the policy's value and compare it to the estimated value of the behavior policy.
If the interval is narrow and lies above the behavior policy's value, deployment is supported by the data.
If the interval is wide or overlaps the behavior policy's value, the improvement is uncertain and the practitioner should either collect more data or accept the behavior policy.

Throughout this process, check for misspecification using the Bellman residual diagnostic of \S\ref{subsec:misspec}.
If residuals are small, the function class is adequate and the pessimistic guarantees apply.
If residuals are large or grow with the stage index, the function class is misspecified, and the practitioner should enrich the features, switch to a model-based approach, or use conservative Q-learning~\citep{kumar2020cql} as a heuristic that is more tolerant of approximation error (though without the same formal guarantees).

The deployment gate of Algorithm~\ref{alg:coveragegate} and the policy certificates of Definition~\ref{def:cert} formalize these checks.
Use them as sequential filters: the coverage gate determines whether to attempt policy improvement at all, and the certificate gate determines whether the improved policy is good enough to deploy.

\section{Partial Observability and Confounding}
\label{sec:pomdp}

The preceding sections assume that the agent observes the true state of the environment at every step.
In many real problems this assumption fails.
A medical decision-support system observes lab results and symptoms, not the underlying disease state.
A dialogue agent observes user utterances, not the user's intent or satisfaction.
A robot navigating from camera images observes pixels, not its precise position and orientation.
When the true state is hidden behind a noisy or incomplete observation channel, the problem becomes a partially observable MDP (POMDP), and the standard MDP theory no longer applies without modification.

POMDPs are, in full generality, computationally and statistically much harder than MDPs.
Planning in a general POMDP is PSPACE-complete~\citep{papadimitriou1987complexity}, and learning without structural assumptions requires sample complexity that scales with the space of possible belief states, which is exponentially large.
The PAC RL literature has therefore focused on structured subclasses where partial observability can be tamed: settings where the observations, despite being high-dimensional or noisy, contain enough information to recover (or approximate) a compact latent state.

This section is deliberately brief.
Our survey focuses on PAC guarantees under the MDP framework, and partial observability is a neighboring territory that we cover only where it intersects directly with the themes of coverage, structure, and function approximation developed in earlier sections.
A full treatment of POMDP learning theory would require a separate survey.

\subsection{Identifiable latent-state models}

The most tractable class of POMDPs for PAC analysis is the one where observations admit a decodable latent representation.
This connects directly to the Block MDP framework discussed in \S\ref{sec:rich}: if the observation space $\cX$ is large but the dynamics depend on a small latent state $z \in [m]$, and if the mapping from observations to latent states can be learned from data, then the agent can plan in the latent MDP with $m$ states rather than in the full observation space.

Under separation and mixing conditions that ensure different latent states produce distinguishable observation distributions, algorithms such as HOMER~\citep{misra2020homer} achieve sample complexity polynomial in $(m, A, H, 1/\varepsilon)$.
The key requirement is identifiability: the observation distribution must vary enough across latent states that a decoder can be learned with controlled error.
When identifiability holds, partial observability adds a representation-learning cost (finding the decoder) but does not fundamentally change the scaling of PAC guarantees relative to a tabular MDP of the same latent size.

Through the CSO lens, identifiable latent-state POMDPs have $\mathsf{Comp} = \mathrm{poly}(m) \cdot A$ (the same as Block MDPs in \S\ref{sec:rich}) and $\mathsf{Cov} = 1$ under online access.
The partial observability is absorbed into the structural term: $m$ replaces $S$, and the cost of learning the decoder is folded into the polynomial dependence on $m$.

\subsection{Low-rank observability}

Beyond exact decoding, low-rank transition factorizations provide a softer form of latent structure.
If the transition operator admits a rank-$r$ decomposition (Definition~\ref{def:lowrank-mdp}), representation learning can exploit this factorization without requiring a strict mapping $\psi : \cX \to [m]$ from observations to discrete latent states.

The guarantees from \S\ref{sec:rich} (Theorem~\ref{thm:block-lowrank}) apply in this setting: sample complexity scales with $\mathrm{poly}(r, H)$ rather than with $|\cX|$, and the algorithms learn the low-rank features from data.
The advantage over Block MDPs is flexibility.
Low-rank structure accommodates continuous latent spaces and soft clustering of observations, while Block MDPs require discrete latent states with hard boundaries.
The cost of this flexibility is that low-rank guarantees typically have less explicit dependence on problem parameters and may involve higher-degree polynomials in $r$ and $H$.

For a practitioner, the choice between Block MDP and low-rank formulations depends on domain knowledge.
If there is reason to believe the environment has a small number of distinct regimes (e.g., disease stages, traffic modes, market conditions), Block MDP assumptions are appropriate and yield tighter bounds.
If the latent structure is continuous or the number of regimes is unclear, low-rank assumptions are safer and more general.

\subsection{Confounded off-policy evaluation}

A different form of partial observability arises in offline settings when the data-generating process involves unobserved confounders.
In a clinical dataset, for example, a doctor's treatment choice may depend on patient features that are recorded in the chart (observed) and on the doctor's clinical intuition or private information (unobserved).
If the unobserved factors influence both the treatment and the outcome, standard OPE estimators produce biased estimates of the target policy's value, because they cannot distinguish the treatment's causal effect from the confounder's influence.

Shi, Zhou, and Gu~\citep{shi2022pomdpOPE} studied OPE in confounded POMDPs and showed that, under structural assumptions (proxy variables, instrumental-variable-style conditions), valid value intervals can be constructed despite the confounding.
The resulting intervals are wider than those from unconfounded OPE (reflecting the additional uncertainty from unobserved variables), but they maintain finite-sample coverage guarantees.

Sensitivity analysis provides a complementary approach~\citep{kallus2021confounding}.
Rather than assuming a specific confounding structure, the practitioner specifies a bound on how much the unobserved confounders can influence the treatment-outcome relationship.
Given this sensitivity parameter, the analysis produces a range of possible policy values, and the width of the range quantifies how much the confounding could distort the evaluation.
When the range is narrow, the OPE estimate is trustworthy despite the confounding; when it is wide, the unobserved variables could be driving the apparent treatment effect, and the practitioner should not trust the evaluation without collecting additional data that resolves the confounding.

\subsection{What partial observability means for deployment}

Two practical lessons emerge from this section.

First, partial observability is not an insurmountable barrier when identifiable latent structure exists.
If the observation space is high-dimensional but a compact latent state can be recovered (through decoder learning or low-rank representation), PAC guarantees carry over from the MDP setting with $m$ or $r$ replacing $S$.
The diagnostics from \S\ref{sec:rich} (spectral analysis, linear Q-heads, contrastive decoding) apply here as well: use them to assess whether latent structure is present and recoverable before committing to a POMDP algorithm.

Second, in offline settings with potential confounding, point estimates of policy value should be treated with skepticism.
Prefer interval estimates that account for the possible influence of unobserved variables, and calibrate the width of the interval using sensitivity analysis.
If the interval is narrow enough to support a confident decision (e.g., the lower bound exceeds the behavior policy's value by a meaningful margin), deployment may be warranted.
If the interval is wide, abstain from policy improvement and focus on collecting data that can resolve the confounding, for instance by recording previously unobserved variables or by designing prospective studies with randomization.

\section{PAC-Bayes Bounds for Reinforcement Learning}
\label{sec:pacbayes}

Every guarantee surveyed so far assumes that the learner commits to a single policy (or value function) and asks whether that policy is near-optimal.
PAC-Bayes theory takes a different approach.
Instead of certifying a single hypothesis, it certifies a distribution over hypotheses, bounding the expected loss of a randomized predictor drawn from a learned posterior $Q$ relative to a fixed prior $P$.
The complexity of the posterior is measured by the Kullback-Leibler divergence $\mathrm{KL}(Q \| P)$: posteriors that stay close to the prior pay a small complexity penalty, while posteriors that deviate substantially must justify the deviation with strong empirical evidence.

PAC-Bayes bounds originated in supervised learning, where they provide some of the tightest known generalization guarantees for neural networks and other overparameterized models.
Their appeal is that they are data-dependent (the bound tightens as the empirical loss decreases), distribution-free (no assumption on the data-generating process beyond i.i.d.\ sampling), and algorithm-independent (any procedure that produces a posterior $Q$ gets a guarantee).
Adapting these bounds to RL introduces complications that do not arise in the supervised setting: the data are not i.i.d.\ (trajectories are correlated within episodes), the loss function (negative return) depends on the policy through the state distribution, and the variance of return estimators can be large.

\subsection{The generic PAC-Bayes bound for policy evaluation}

The simplest application of PAC-Bayes to RL is policy evaluation: given a dataset of $n$ i.i.d.\ trajectories collected under a behavior policy, bound the expected return of policies drawn from a posterior $Q$.

Let $\Pi$ be a policy class, $P$ a prior distribution on $\Pi$ chosen before observing data, and $\hat{R}(\pi)$ an empirical return estimator (for instance, an importance-weighted average of observed returns under $\pi$).
Fard, Pineau, and Szepesvári~\citep{fard2012pacbayes} established the basic PAC-Bayes inequality for this setting: with probability at least $1 - \delta$, for all posteriors $Q$ on $\Pi$ simultaneously,
\[
\E_{\pi \sim Q}\big[R(\pi)\big]
\;\le\;
\E_{\pi \sim Q}\big[\hat{R}(\pi)\big]
\;+\;
\sqrt{\frac{\mathrm{KL}(Q \| P) + \ln(1/\delta)}{2n}}
\;+\; \text{bias and variance corrections}.
\]
The first term is the empirical expected return under the posterior.
The second term is the complexity penalty: it grows with the divergence between $Q$ and $P$ and shrinks with more data.
The third term absorbs the additional error from estimating returns in the RL setting (importance weighting variance, horizon-dependent bias, correlation within episodes).

The bound holds uniformly over all posteriors $Q$, which means the learner can optimize $Q$ after seeing the data (to minimize the right-hand side) and the guarantee still holds.
This is the key advantage over single-policy PAC bounds: the learner does not need to commit to a policy class or realizability assumption in advance.
Any posterior that achieves low empirical loss and stays close to the prior gets a valid certificate.

\subsection{Beyond standard bounds}

The basic PAC-Bayes inequality assumes bounded losses and i.i.d.\ data, both of which are violated in typical RL settings.
Rivasplata et al.~\citep{rivasplata2020pacbayes} extended the framework to handle unbounded losses, which arise naturally in RL when returns can have high variance (e.g., in long-horizon or sparse-reward problems).
Their analysis replaces the bounded-loss assumption with moment conditions, yielding bounds that degrade gracefully as the return variance increases rather than becoming vacuous.

Flynn et al.~\citep{flynn2023pacbayesbandit} surveyed PAC-Bayes bounds for bandit problems (the single-stage special case of RL) and provided experimental comparisons showing that PAC-Bayes bounds can be substantially tighter than classical uniform-convergence bounds when the posterior concentrates on a small region of policy space.
Their findings suggest that the PAC-Bayes approach is most valuable when the policy class is large (making uniform bounds loose) but the good policies cluster together (making a concentrated posterior cheap in KL terms).

Tasdighi et al.~\citep{tasdighi2024pbac} pushed the framework further into sequential settings with deep exploration, using PAC-Bayes bounds to regularize policy search in continuous-control tasks.
Their approach treats the PAC-Bayes bound as an objective function: instead of optimizing expected return and then certifying the result, the algorithm directly minimizes the PAC-Bayes upper bound on the true return, jointly optimizing the posterior and the empirical performance.
This produces policies with built-in generalization certificates, though the resulting optimization problem is harder than standard policy gradient methods.

\subsection{How PAC-Bayes complements the rest of this survey}

PAC-Bayes occupies a distinct niche in the PAC RL toolkit, and understanding where it fits clarifies when to use it.

The guarantees in \S\ref{sec:tabular} through \S\ref{sec:offline} are model-based or value-based: they assume a function class $\cF$, require realizability ($Q^\star \in \cF$) or completeness ($\mathcal{T}_h \cF \subseteq \cF$), and derive sample complexity bounds that scale with the capacity of $\cF$.
If the assumptions hold, these bounds are tight and the algorithms are efficient.
If the assumptions fail (the function class is misspecified, completeness is violated), the bounds become vacuous and the algorithms may produce bad policies without warning.

PAC-Bayes bounds trade tightness for robustness to misspecification.
They do not require realizability: even if no single policy in $\Pi$ is optimal, the bound certifies the expected performance of the posterior $Q$, which may average over many good-enough policies.
They do not require Bellman completeness: the bound operates on the return directly, bypassing the Bellman equation entirely.
The cost of this generality is that PAC-Bayes bounds are typically looser than model-based bounds when realizability holds, because they do not exploit the sequential structure of the MDP.

This tradeoff suggests a practical division of labor.
When the function class is well-specified and verifiable (e.g., linear features that pass the residual diagnostic of \S\ref{subsec:misspec}), use the model-based or value-based guarantees from earlier sections, which are tighter.
When the function class is uncertain or likely misspecified (e.g., a neural network policy in a complex environment), PAC-Bayes bounds provide a fallback that does not rely on structural assumptions.
In offline RL, PAC-Bayes can complement pessimism: use pessimistic algorithms to learn a candidate policy, then apply a PAC-Bayes bound to certify (or fail to certify) the candidate before deployment.

The prior $P$ in the PAC-Bayes bound is a design choice that encodes the practitioner's inductive bias.
Natural choices in RL include a uniform prior over a discrete policy class, a Gaussian prior centered on the behavior policy (encoding a preference for policies close to the data-generating process), or a prior derived from a pre-trained model.
The tighter the prior matches the region of good policies, the smaller the KL penalty and the tighter the bound.
Choosing the prior well is the main art of applying PAC-Bayes in practice; a poor prior (far from all good policies) can make the bound vacuous even with abundant data.

\subsection{Limitations and open questions}

PAC-Bayes for RL remains less developed than the model-based theory surveyed in earlier sections.

The main technical gap is the treatment of temporal dependence.
Most PAC-Bayes results for RL either assume i.i.d.\ trajectories (which rules out online learning and single-trajectory settings) or absorb temporal correlation into the variance terms, which can make the bounds loose for long horizons.
Extending PAC-Bayes theory to handle within-episode dependence without excessive penalty is an active research direction.

A second gap is computational.
Optimizing the PAC-Bayes bound (finding the posterior $Q$ that minimizes the right-hand side) is a variational problem that requires approximation in practice.
The quality of the bound depends on how well the variational family approximates the true optimal posterior, and standard variational inference may not be adequate for the complex loss landscapes that arise in RL.

Despite these limitations, PAC-Bayes provides a unique capability that no other framework in this survey offers: data-dependent, assumption-light certificates for distributions over policies.
For practitioners working with complex function classes where realizability cannot be verified, this is a valuable fallback.


\section{Synthesis: What the CSO Lens Reveals}
\label{sec:synth}

After traversing the landscape of PAC guarantees across settings, it is worth stepping back to ask what patterns emerge.
This section takes stock of the field through the CSO framework, identifying where the theory is mature, where it is not, and what the decomposition into coverage, structure, and objective tells us about the shape of future progress.

\subsection{Where the theory is settled}

The tabular online regime is essentially resolved.
Zhang et al.~\citep{zhang2024settling} confirmed that the $\tTheta(SAH^3/\varepsilon^2)$ rate is tight, and the uniform-PAC framework~\citep{uniformpac2017} provides a clean bridge to high-probability regret.
In the reward-free tabular setting, Jin et al.~\citep{jin2020rewardfree} established near-optimal dependence on all parameters, including the necessary $S^2$ factor for covering all possible downstream rewards.
For linear MDPs, PAC and regret bounds with polynomial dependence on $(d, H)$ are known~\citep{jin2020linear}, and uniform-PAC extensions exist under Bellman completeness~\citep{he2021uniformpaclinear}.
Offline RL in the tabular model-based regime is settled at the minimax level~\citep{li2024aos}, and pessimistic linear-MDP control with explicit $C_\star$ dependence is well understood~\citep{jin2021pevi,shi2022pql}.

What these settled results share, viewed through CSO, is that each axis is addressed independently and cleanly.
The structural term is known exactly (or up to polynomial degree in $H$).
The coverage term is either trivial ($\mathsf{Cov} = 1$ for online) or characterized by a single coefficient ($C_\star$ for offline).
The objective term is standard PAC or uniform-PAC.
Difficulty arises precisely when these clean separations break down.

\subsection{Where the axes interact}

The hardest open problems are those where multiple CSO axes are coupled, so that improving one factor without addressing the others is insufficient.

Offline RL under misspecification couples coverage and structure.
If the function class is wrong ($Q^\star \notin \cF$), additional data cannot fix the approximation error, and better coverage cannot compensate for systematic bias.
The total error decomposes into approximation, estimation, and coverage components, but these components interact: misspecification changes which states matter for coverage, and poor coverage amplifies the effect of misspecification.
Sharp characterizations of this three-way interaction, with algorithms that are both pessimistic and robust to model error, do not yet exist.

Agnostic low-rank learning couples structure and computation.
The statistical rates for low-rank MDPs under approximate rank-$r$ structure are known~\citep{dann2021agnostic}, but achieving them with polynomial-time algorithms requires additional assumptions (exact realizability, specific rank structure) that limit applicability.
The gap between what is statistically possible and what is computationally achievable remains wide.

Instance-dependent identification with function approximation couples structure and objective.
In the tabular setting, gap-weighted rates replace the worst-case $SA$ factor with a sum over inverse squared gaps weighted by reachability~\citep{wagenmaker2022idpac,tirinzoni2023optimistic}.
Extending these results to linear or low-rank models requires new tools for measuring reachability in feature space, where the notion of "gap at a state" must be replaced by something that accounts for feature geometry.

Kernel and NTK uniform-PAC guarantees couple structure and verifiability.
Current results require Bellman completeness, which is a condition on the interaction between the function class and the MDP dynamics.
For linear classes, completeness can be checked via residual diagnostics (\S\ref{subsec:misspec}).
For kernel classes, no practical verification procedure exists, and developing one is tied to understanding the spectral structure of Bellman operators in RKHS.

\subsection{What CSO does and does not capture}

The CSO decomposition has proved useful as an organizing tool throughout this survey, but it is worth stating clearly what it misses.

It does not capture computational complexity.
Two settings with identical CSO coordinates can differ dramatically in whether polynomial-time algorithms exist.
Bilinear classes and general finite-$d_{\mathrm{BE}}$ classes have similar statistical rates, but the former admit efficient algorithms while the latter may require exponential-time search.

It abstracts away horizon exponents.
The template writes $\mathrm{poly}(H)$ as a single factor, but $H^3$ vs.\ $H^4$ vs.\ $H^6$ matters in practice, and the exact exponent often reflects deep structural differences (independent vs.\ correlated estimation errors, stage-wise vs.\ global confidence sets).

It does not apply to objectives outside the PAC family.
Bayesian regret, simple regret without the PAC wrapper, and constrained optimization (where the agent must satisfy safety constraints while optimizing reward) all fall outside the CSO template.

These limitations do not invalidate the framework, but they circumscribe its scope.
CSO is a reading tool for the PAC RL literature of 2018 to 2025, not a universal theory of reinforcement learning.

\subsection{A practitioner's takeaway}

For applied researchers, the synthesis reduces to a short checklist.

Identify your access mode (online, offline, reward-free) to fix the coverage axis.
Characterize your function class (tabular, linear, kernel, low-rank, neural) to fix the structure axis.
Specify your objective (one good policy, uniform guarantees, policy evaluation, identification) to fix the objective axis.
Look up the rate in Table~\ref{tab:taxonomy} and the relevant section.
Before trusting the guarantee, verify the underlying assumptions: run the Bellman residual test (Algorithm~\ref{alg:bellman}) for realizability and completeness, estimate coverage proxies (Algorithm~\ref{alg:coveragegate}) for offline data adequacy, and gate deployment on policy certificates (Definition~\ref{def:cert}).
If the assumptions fail or the bound is vacuous, the CSO parsing tells you which axis is responsible and suggests a remedy: better features for the structure axis, more diverse data for the coverage axis, or a relaxed objective (evaluation instead of control) for the objective axis.

\section{Open Problems}
\label{sec:open}

The settled results cataloged in \S\ref{sec:synth} leave substantial territory unexplored.
This section maps that territory, distinguishing problems that appear solvable with current techniques from those that require new ideas.
The organizing principle is the CSO decomposition: the most tractable open problems involve a single axis, while the hardest involve interactions between axes that current theory cannot disentangle.

\subsection{Problems within reach}

Three open problems seem approachable with extensions of existing tools, and we would expect significant progress within the next few years.

The first is uniform-PAC guarantees for kernel and RKHS value classes under conditions that a practitioner can verify.
Current kernel RL results (Theorem~\ref{thm:kernel}) achieve rates scaling with the effective dimension $d_{\mathrm{eff}}(\lambda)$, but they assume Bellman completeness, which is a condition on the joint behavior of the function class and the MDP dynamics.
For linear classes, completeness can be checked empirically using the residual diagnostic of \S\ref{subsec:misspec}.
For kernel classes, no analogous test exists.
The target is an algorithm achieving $N = \tO(d_{\mathrm{eff}}(\lambda) \cdot \mathrm{poly}(H) \cdot \varepsilon^{-2})$ under online access, with completeness verified (or bypassed) by a data-driven residual test rather than assumed.
The technical challenge is that kernel function classes are infinite-dimensional, so residual tests must control approximation error in RKHS norm rather than in a finite parameter count.
Recent work on spectral methods for Bellman operators suggests that eigenvalue thresholding could provide a practical proxy, but a complete solution with finite-sample guarantees remains open.

The second is instance-dependent identification with function approximation.
In the tabular setting, Wagenmaker et al.~\citep{wagenmaker2022idpac} and Tirinzoni et al.~\citep{tirinzoni2023optimistic} showed that gap-weighted rates can replace the worst-case $SA$ factor, achieving sample complexity proportional to $\sum_{h,s} q_h(s) \sum_a \Delta_h(s,a)^{-2}$ (Theorem~\ref{thm:tab-id}).
Extending this to linear or low-rank models is natural but not straightforward.
The difficulty is that gaps in the tabular setting are defined state by state, while in the function approximation setting, the relevant notion of ``gap'' must account for feature geometry: two state-action pairs with similar features cannot have independently estimated gaps, and the effective gap is a property of directions in feature space rather than individual states.
A clean formulation would define gaps in terms of the eigenstructure of the design matrix and derive instance-dependent rates that recover the tabular bounds when features are one-hot indicators.

The third is scalable coverage estimation with finite-sample guarantees for offline RL deployment.
The coverage diagnostics in \S\ref{subsec:coverage} (density ratios, ridge leverage scores, effective sample sizes) are practically useful but lack rigorous finite-sample confidence intervals in the function approximation setting.
The target is an estimator $\widehat{C}_\star$ with a confidence interval $[\widehat{C}_\star \pm \Delta]$ such that the deployment gate of Algorithm~\ref{alg:coveragegate} can make provably correct accept/reject decisions with controlled Type I and Type II error rates.
The main obstacle is that coverage estimation requires estimating density ratios in high-dimensional spaces, and the minimax rates for density ratio estimation under smoothness or structural constraints are not fully characterized.

\subsection{Frontier problems}

Three problems appear to require fundamentally new techniques or insights, and progress on them would represent a qualitative advance in the field.

The first frontier problem is agnostic low-rank and latent-state PAC learning.
Current low-rank MDP guarantees (Theorem~\ref{thm:block-lowrank}) assume exact rank-$r$ structure or exact decodability.
When these assumptions hold approximately (the transition operator has rank $r + \Delta r$, or the decoder has error $\varepsilon_{\mathrm{decode}}$), the existing algorithms incur an additive misspecification penalty that does not vanish with more data~\citep{dann2021agnostic}.
The target is rates polynomial in $(r, H, 1/\varepsilon)$ that degrade gracefully in the misspecification level $\varepsilon_{\mathrm{approx}}$, achieved by oracle-efficient algorithms.
The difficulty is twofold.
Statistically, the algorithm must estimate both the representation (features or decoder) and the value function simultaneously, and errors in representation learning feed into value estimation in ways that are hard to decouple.
Computationally, joint optimization over representations and policies is nonconvex, and it is unclear whether polynomial-time algorithms can match the statistical rates even in benign cases.

The second is offline RL under simultaneous misspecification and partial coverage.
As discussed in \S\ref{sec:offline}, the total error in offline RL decomposes into approximation error (from misspecification), estimation error (from finite data), and coverage error (from distribution mismatch).
Current theory handles each component separately: pessimistic algorithms control estimation and coverage error under exact realizability, and agnostic bounds control approximation error under good coverage.
The target is a sharp characterization of the three-way tradeoff, specifying how approximation error $\varepsilon_{\mathrm{approx}}$, sample size $n$, and concentrability $C_\star$ jointly determine the achievable policy quality, together with tractable pessimistic algorithms that are simultaneously robust to model error and adaptive to coverage quality.
This problem is at the frontier because the interaction between misspecification and coverage is not merely additive: misspecification changes which states matter for the optimal policy, which in turn changes what coverage is needed, creating a circular dependency that linear decompositions cannot capture.

The third is information-theoretic lower bounds for kernel and NTK reinforcement learning.
Upper bounds for kernel RL scale with $d_{\mathrm{eff}}(\lambda)$ (Theorem~\ref{thm:kernel}), but matching lower bounds that confirm $d_{\mathrm{eff}}$ as the right complexity measure are largely absent.
Without lower bounds, it is unclear whether the $d_{\mathrm{eff}}$-based rates are tight or whether tighter characterizations (perhaps involving finer spectral quantities) are possible.
The target is a lower bound of $\Omega(d_{\mathrm{eff}}(\lambda) \cdot \mathrm{poly}(H) \cdot \varepsilon^{-2})$ for kernel RL under appropriate conditions, together with a resolution of the optimal horizon exponent (which currently ranges from $H^4$ to $H^6$ in the upper bounds).
The difficulty is that lower bound constructions for kernel problems require building hard MDP instances within the RKHS, and the smoothness constraints imposed by the kernel make it challenging to embed the combinatorial gadgets that standard lower bound techniques rely on.

\subsection{What is known to be impossible}

Not every desirable guarantee is achievable, and it is important to distinguish open problems (where progress is possible) from impossibility results (where the difficulty is inherent).

Offline RL without coverage assumptions is one such barrier.
Without a bound on the concentrability coefficient $C_\star$ or a similar quantity, consistent policy improvement over the behavior policy is impossible from offline data alone.
The reason is fundamental: if the behavior policy never visits a state that the optimal policy would, the dataset contains exactly zero information about what to do there, and no algorithm can extrapolate reliably to unseen states without structural assumptions.
In this regime, the right object is not a near-optimal policy but an interval estimate of policy value (as in OPE), which honestly quantifies the uncertainty due to missing coverage rather than pretending it does not exist.

Kernel RL without Bellman completeness faces a related barrier.
In general contextual decision processes, RKHS value classes that are not closed under Bellman updates can exhibit aliasing: two value functions that agree on the observed data but disagree on unseen states in a way that backward induction cannot detect.
This aliasing prevents PAC guarantees without additional structure (such as a generative model that allows querying arbitrary states) or a verified bound on the Bellman residual.
The completeness condition is therefore not merely a technical convenience but a necessity for the type of guarantees this survey considers.

Agnostic function approximation without computational structure is the third barrier.
With unrestricted function classes and no realizability assumption, polynomial sample complexity can be achieved information-theoretically (through exhaustive search over $\cF$), but polynomial-time algorithms are not known and are unlikely to exist in full generality.
Computational hardness results from the supervised learning literature (e.g., hardness of agnostic learning of halfspaces) suggest that similar barriers apply in the RL setting, and that progress on agnostic PAC RL will require either restricting the function class (as in the bilinear or low-rank settings) or accepting super-polynomial computation.

\section{Practical Toolkit: Coverage, Diagnostics, and Deployment}
\label{sec:practice}

The theoretical results in this survey are only useful if a practitioner can determine whether they apply to a given problem and, if so, whether the data at hand are sufficient for deployment.
This section provides concrete procedures for making those determinations.
We address three questions in sequence: does the function class fit the problem (misspecification diagnostics), does the data adequately support the target policy (coverage estimation), and is the learned policy good enough to deploy (deployment gates and certificates)?

The procedures here are not new theorems but practical operationalizations of the theory developed in earlier sections.
They are designed to be run before and after offline RL pipelines, before deploying online-learned policies, and as ongoing monitors during deployment.

\subsection{Diagnosing misspecification}
\label{subsec:misspec}

The guarantees of \S\ref{sec:function} assume that the value function class $\cF$ contains (or closely approximates) the true optimal value function, and that $\cF$ is closed under Bellman updates.
When either assumption fails, the guarantees do not hold, and algorithms that rely on them can produce confidently wrong policies.
Checking these assumptions before invoking any bound is therefore the first step in any rigorous RL pipeline.

The Bellman residual test provides a simple, practical diagnostic.
The idea is to fit value functions using the candidate function class and check whether the fitted functions satisfy the Bellman equation on held-out data.
If they do, the function class is consistent with the MDP dynamics and the guarantees are likely to apply.
If they do not, the function class is misspecified and the practitioner should either enrich the features, switch to a more flexible model class, or fall back to methods that do not require realizability (such as PAC-Bayes bounds from \S\ref{sec:pacbayes}).

\begin{algorithm}[H]
\caption{BellmanResidualTest$(\phi, \mathsf{Alg}, \varepsilon)$}
\label{alg:bellman}
\begin{algorithmic}[1]
\State Collect $n$ episodes under a random (or diverse) policy.
\State Split data into training and held-out sets.
\State For each stage $h = H, H{-}1, \ldots, 1$: fit $f_h(s,a) = \phi(s,a)^\top w_h$ by ridge regression on training data, using $r_h + \max_{a'} f_{h+1}(s', a')$ as the regression target.
\State Compute held-out Bellman residuals: $\mathcal{E}_{h,i} = |f_h(s_i, a_i) - (r_{h,i} + \max_{a'} f_{h+1}(s'_i, a'))|$ for each held-out transition.
\State Report the mean squared residual $\bar{\mathcal{E}}_h^2 = \frac{1}{n_{\mathrm{held}}} \sum_i \mathcal{E}_{h,i}^2$ at each stage.
\State \textbf{Pass} if $\bar{\mathcal{E}}_h^2 \le \varepsilon^2$ for all $h$; otherwise flag misspecification.
\end{algorithmic}
\end{algorithm}

The interpretation of the test output requires some care.
Small residuals at all stages indicate that the function class is approximately realizable and approximately Bellman-complete: the fitted value functions nearly satisfy the Bellman equation, so backward induction within $\cF$ is well-behaved.
Residuals that grow with decreasing $h$ (i.e., residuals are small at later stages but large at early stages) indicate that Bellman completeness fails: the function class can represent terminal values but not the results of iterated Bellman backups, and errors compound as backward induction proceeds.
Residuals that are uniformly large suggest that realizability itself fails: $Q^\star$ is not well-approximated by any function in $\cF$, and a richer function class is needed.

When the test flags misspecification, several remedies are available depending on the pattern of residuals.
If residuals are large only at specific stages or in specific regions of the state space, adding stage-dependent features or nonlinear basis functions (polynomial terms, radial basis functions, interaction terms) may resolve the issue.
If residuals are uniformly large, the problem may require kernel methods (\S\ref{sec:function}), learned representations (\S\ref{sec:rich}), or an entirely different modeling approach.
If residuals are small on training data but large on held-out data, the function class may be overparameterized, and regularization or a simpler model is appropriate.

\subsection{Estimating coverage for offline data}
\label{subsec:coverage}

Once the function class has passed the misspecification diagnostic, the next question for offline RL is whether the dataset $\cD$ provides adequate support for the target policy.
The theory of \S\ref{sec:offline} shows that the concentrability coefficient $C_\star$ governs offline sample complexity, but $C_\star$ depends on the unknown optimal policy and cannot be computed exactly.
Three complementary diagnostics provide approximate answers.

The most direct approach is density-ratio estimation.
For each stage $h$, fit a classifier to distinguish state-action pairs drawn from the dataset (behavior policy $\mu$) from state-action pairs drawn from rollouts of the candidate policy $\pi$ under a learned model.
The fitted classifier probabilities yield importance weight estimates $\hat{w}_h(x,a) \approx d_h^\pi(x,a) / \mu_h(x,a)$, which should be clipped at a maximum value $w_{\max}$ to stabilize the estimates.
From the clipped weights, compute the effective sample size $\mathrm{ESS}_h = (\sum_i \hat{w}_{h,i})^2 / \sum_i \hat{w}_{h,i}^2$, the 99th percentile of the weight distribution $w_{h,0.99}$, and the maximum weight $\max_i \hat{w}_{h,i}$.
The effective sample size measures how many ``unweighted'' samples the importance-weighted dataset is worth: an ESS of 500 from a dataset of 10{,}000 means that 95\% of the data is irrelevant to the target policy, while an ESS close to the dataset size means the behavior and target policies have similar support.

A complementary diagnostic uses ridge leverage scores, which measure coverage in feature space rather than in state-action space.
For linear or kernel value estimation, compute the stagewise design matrix $X_h$ from the dataset and evaluate the leverage score $\tau_{h,i} = x_{h,i}^\top (X_h^\top X_h + \lambda I)^{-1} x_{h,i}$ for each data point.
The sum of leverage scores equals the effective dimension $d_{\mathrm{eff}}(\lambda)$, and the distribution of individual scores reveals where coverage is strong (low leverage, well-represented direction) and where it is weak (high leverage, poorly represented direction).
If many data points have leverage scores near $1$, the design matrix is poorly conditioned and the ridge regression estimates in those directions are unreliable.

A third diagnostic, more qualitative but often revealing, is visual inspection of state-action occupancy.
Bin the state-action pairs (or their embeddings) into a grid and plot occupancy heatmaps across stages.
Regions where the behavior policy visits frequently appear as bright areas; regions where the target policy would visit but the behavior policy does not appear as dark gaps.
Large dark gaps in reward-relevant regions of the state space indicate that coverage is insufficient and that pessimistic algorithms will produce overly conservative policies or that policy improvement should not be attempted.

\begin{algorithm}[H]
\caption{CoverageGate$(\cD, \pi, \lambda, w_{\max})$}
\label{alg:coveragegate}
\begin{algorithmic}[1]
\State \textbf{for} $h = 1 \ldots H$ \textbf{do}
\State \quad Fit logistic regression classifier to distinguish $(x_h, a_h) \sim \cD$ vs.\ $(x_h, a_h) \sim$ rollouts of $\pi$ under a learned model; obtain importance weight estimates $\hat{w}_h$.
\State \quad Clip: $\hat{w}_h \leftarrow \min(\hat{w}_h, w_{\max})$.
\State \quad Compute $\mathrm{ESS}_h$, $q_{0.99}(\hat{w}_h)$, $\max \hat{w}_h$.
\State \quad Compute ridge leverage scores from $X_h$ and $\lambda$; summarize $\bar{\tau}_h$, $q_{0.95}(\tau_h)$.
\State \textbf{end for}
\State \textbf{return} \textsc{Pass} if $\min_h \mathrm{ESS}_h \ge 200$ and $q_{0.99}(\hat{w}_h) \le 30$ and $q_{0.95}(\tau_h) \le 0.5$; else \textsc{Warn/Abstain}.
\end{algorithmic}
\end{algorithm}

The thresholds in Algorithm~\ref{alg:coveragegate} (ESS $\ge 200$, $q_{0.99} \le 30$, $q_{0.95}(\tau) \le 0.5$) are not sharp theoretical cutoffs but practical defaults calibrated to avoid the most common failure modes.
An ESS below 200 means that fewer than 200 effective samples inform the importance-weighted estimate, which is typically insufficient for reliable policy evaluation at moderate $\varepsilon$.
A 99th percentile weight above 30 means that a few data points dominate the importance-weighted estimate, making it sensitive to noise in those points.
A 95th percentile leverage score above 0.5 means that many data points contribute almost as much information as a single unconstrained observation, indicating weak coverage in the corresponding feature directions.
Practitioners should treat these thresholds as starting points and adjust them based on the stakes of the deployment: higher-stakes applications warrant more conservative thresholds.

One implementation detail deserves emphasis: use cross-fitting for all nuisance models (behavior policy estimates, learned dynamics, density ratio classifiers).
Fitting the nuisance model on the same data used to evaluate coverage introduces optimistic bias, because the model overfits to the evaluation sample.
Cross-fitting (fitting on one fold, evaluating on another, and averaging across folds) eliminates this bias at negligible computational cost.
For kernel-based pipelines where leverage score computation is expensive ($O(n^3)$ for exact kernel matrices), Nyström approximations provide $O(n k^2)$ alternatives with $k$ landmark points, trading a small approximation error for substantial speedup.

\subsection{Deployment gates}
\label{subsec:gates}

The misspecification diagnostic and coverage estimator feed into a two-stage deployment gate that determines whether a learned policy should be deployed.

The first stage is the coverage gate (Algorithm~\ref{alg:coveragegate}).
If coverage is inadequate (the gate returns \textsc{Warn} or \textsc{Abstain}), the practitioner should not attempt policy improvement from the offline data.
Instead, use off-policy evaluation (\S\ref{sec:offline}) to estimate the value of candidate policies without deploying them, and report interval estimates rather than point improvements.
If additional data collection is possible, guide it toward the under-covered regions identified by the leverage scores and occupancy heatmaps.

The second stage is the certificate gate.
Even when coverage is adequate, the learned policy should be certified before deployment using the per-episode policy certificates of Definition~\ref{def:cert}.
The certificate $U_t$ provides a data-dependent upper bound on the suboptimality of the current policy: deploy only if $U_t \le \varepsilon$ for the desired tolerance.
If $U_t$ remains large despite adequate coverage, the algorithm may need more data, a better function class, or a different optimization procedure.
The certificate is not a one-time check but an ongoing monitor: in online settings, track $U_t$ across episodes and halt deployment if it rises above the tolerance, which can indicate distribution shift or model degradation.

The two gates serve different purposes and catch different failure modes.
The coverage gate catches data inadequacy before the algorithm runs, preventing wasted computation on problems where the data cannot support improvement.
The certificate gate catches algorithmic failure after the algorithm runs, preventing deployment of policies that looked good during training but do not generalize.
Running both gates in sequence provides layered protection against the two most common deployment failures in offline RL: overconfident improvement from narrow data, and overfitting to the training distribution.

\subsection{A worked example}
\label{subsec:numeric}

To make the diagnostic workflow concrete, consider a $10 \times 10$ gridworld with horizon $H = 30$.
A random behavior policy $\mu$ (uniform over actions at every state) generates 5{,}000 episodes of offline data.
An offline RL algorithm (pessimistic Q-learning with tabular features) produces a candidate policy $\pi$.

Running Algorithm~\ref{alg:coveragegate} on this setup yields $\min_h \mathrm{ESS}_h = 74$, $q_{0.99}(\hat{w}_h) = 115$, and $q_{0.95}(\tau_h) = 0.81$.
All three metrics fail their thresholds.
The low effective sample size indicates that the candidate policy $\pi$ visits regions of the gridworld that the random behavior policy rarely reaches, which is expected: a trained policy concentrates on high-reward paths while a random policy spreads mass uniformly, including over low-reward regions that the trained policy avoids.
The high weight percentile confirms that a few state-action pairs carry disproportionate importance, making the value estimate fragile.
The high leverage scores indicate poor coverage in certain feature directions.
The gate returns \textsc{Abstain}: this dataset does not support confident deployment of $\pi$.

The practitioner then collects 200 additional online episodes, using the pessimistic policy $\pi$ with $\varepsilon$-greedy perturbation to explore near the regions where coverage was weak.
Running the coverage gate again yields $\min_h \mathrm{ESS}_h = 312$, $q_{0.99}(\hat{w}_h) = 21$, and $q_{0.95}(\tau_h) = 0.42$.
All three metrics now pass.
The targeted data collection filled the coverage gaps: the effective sample size tripled, the weight tail shrank by a factor of five, and the leverage scores dropped below the threshold.
The gate returns \textsc{Pass}, and the practitioner proceeds to the certificate check.

Computing the policy certificate yields $U_t = 0.12$ (in units of maximum per-episode return), which is below a deployment tolerance of $\varepsilon = 0.15$.
Both gates pass, and the policy is cleared for deployment.

This example illustrates two practical points.
The initial dataset, despite containing 5{,}000 episodes, was insufficient because coverage quality matters more than dataset size.
And targeted supplementary data collection (guided by the coverage diagnostic) was far more efficient than collecting another 5{,}000 random episodes would have been: 200 targeted episodes achieved what thousands of random episodes could not.

\section{Ethics Statement}\label{sec:impact}
This survey aggregates theory rather than proposing new algorithms, but its claims can influence practice in safety-critical domains such as healthcare, robotics, and education.
The primary ethical risk is misapplication of guarantees outside their assumptions, for example invoking linear-MDP bounds with misspecified features or deploying an offline policy under poor coverage, either of which can harm users.
Mitigations emphasized throughout the survey include explicit assumption checklists covering realizability, horizon scaling, and coverage proxies; per-episode policy certificates to gate deployment; abstention and interval-valued off-policy evaluation when coverage is weak; and reporting structural and coverage diagnostics alongside results.
Fairness risks arise when offline datasets encode historical bias; coverage analysis and sensitivity checks can reveal regions where performance is unreliable.
We recommend conservative defaults (pessimism, OPE before improvement), public documentation of assumptions, and auditing procedures that include the failure modes cataloged in this survey.

\section*{Reproducibility Statement}\label{sec:repro}
This survey covers the time window 2018--2025 and includes peer-reviewed results from JMLR, PMLR venues (COLT, ALT, ICML), and NeurIPS, supplemented by widely cited arXiv preprints when they anchor common bounds; pre-2018 works are cited for foundations.
All results are stated using unified MDP, POMDP, and CDP notation introduced in \S\ref{sec:preliminaries}, with a consolidated glossary in Appendix~\ref{app:notation} and first-use definitions for every parameter ($H, S, A, d, r, B, W, d_{\mathrm{BE}}, d_{\mathrm{eff}}(\lambda), C_\star$).

Canonical guarantees are restated with explicit parameter dependencies throughout; rate tables flag where logarithmic factors are suppressed, and primary sources are cited for exact constants and exponents.
The literature search used queries including \emph{uniform-PAC reinforcement learning}, \emph{PAC RL finite horizon}, \emph{Bellman rank}, \emph{Bellman-Eluder dimension}, \emph{linear MDP}, \emph{low-rank MDP / Block MDP}, \emph{reward-free exploration}, \emph{offline RL pessimism / concentrability}, and \emph{PAC-Bayes reinforcement learning}, scanning JMLR, PMLR (COLT/ALT/ICML), NeurIPS, and selected arXiv preprints for works later published at these venues; we favored papers with clear assumptions and finite-sample theorems.

No new datasets or code are introduced; all figures are conceptual, and canonical implementations are referenced where applicable.
When rates differ across papers (e.g., horizon exponents), we state the dependency class and cite the precise theorem for constants; competing viewpoints are mentioned where relevant.
All citations correspond to versions current as of October 2025.

\section*{Limitations}
Our focus is episodic finite--horizon and fixed--confidence (PAC / uniform--PAC) guarantees; we do not cover continuous--time control, average--reward/ergodic settings, or detailed deep--RL practice beyond linear/kernel/NTK regimes. Several summaries suppress polylogarithms; constants and sharp horizon exponents are deferred to primary sources. Our treatment of partial observability is limited to identifiable latent--state / low--rank subclasses and confounded OPE. Many results assume realizability or Bellman completeness; robustness to misspecification is highlighted as an open problem. Finally, while we provide diagnostics and decision rules, we do not release audited toolchains for coverage estimation or certificate computation.



\paragraph{Bibliography organization.}
References are grouped thematically: \emph{Foundations} (Valiant, Bellman, Howard, Puterman, Strehl, Szepesvári, Kaelbling, Sutton--Barto); \emph{Uniform--PAC \& tabular} (Dann--Lattimore--Brunskill; Domingues; Zhang); \emph{Structural complexity} (Jiang; Sun; Jin--Liu--Miryoosefi; Du--Kakade); \emph{Function approximation} (Jin--Yang--Wang; He--Zhou--Gu; Yang et al.); \emph{Rich observations} (Du; Misra; Agarwal; Dann--agnostic); \emph{Reward--free} (Jin; Kaufmann; Wang); \emph{Offline} (Jin--PEVI; Shi--PQL; Li--AOS; Kallus; Jiang); \emph{PAC--Bayes} (Fard; Rivasplata; Flynn; Tasdighi).

\makeatletter
\IfFileExists{tmlr.bst}{\bibliographystyle{tmlr}}{\bibliographystyle{plainnat}}
\makeatother
\bibliography{pac_rl_survey}

@inproceedings{uniformpac2017,
  author    = {Christoph Dann and Tor Lattimore and Emma Brunskill},
  title     = {Unifying {PAC} and Regret: Uniform-{PAC} Bounds for Episodic {RL}},
  booktitle = {NeurIPS},
  year      = {2017},
  url       = {https://arxiv.org/abs/1703.07710}
}

@inproceedings{domingues2021lowerbounds,
  author    = {Omar Darwiche Domingues and Pierre M{\'e}nard and Emilie Kaufmann and Michal Valko},
  title     = {Episodic Reinforcement Learning in Finite {MDP}s: Minimax Lower Bounds Revisited},
  booktitle = {ALT},
  series    = {PMLR},
  volume    = {132},
  pages     = {578--598},
  year      = {2021}
}

@inproceedings{zhang2024settling,
  title     = {Settling the sample complexity of online reinforcement learning},
  author    = {Zhang, Zihan and Chen, Yuxin and Lee, Jason D and Du, Simon S},
  booktitle = {COLT},
  volume    = {247},
  series    = {PMLR},
  year      = {2024}
}

@inproceedings{dann2019certificates,
  author    = {Christoph Dann and Lihong Li and Wei Wei and Emma Brunskill},
  title     = {Policy Certificates: Towards Accountable Reinforcement Learning},
  booktitle = {ICML},
  volume    = {97},
  pages     = {1507--1516},
  year      = {2019}
}

@inproceedings{jiang2017olive,
  author    = {Nan Jiang and Akshay Krishnamurthy and Alekh Agarwal and John Langford and Robert E. Schapire},
  title     = {Contextual Decision Processes with Low Bellman Rank are {PAC}-Learnable},
  booktitle = {ICML},
  volume    = {70},
  year      = {2017}
}

@inproceedings{sun2019witness,
  author    = {Wen Sun and Nan Jiang and Akshay Krishnamurthy and Alekh Agarwal and John Langford},
  title     = {Model-Based RL in Contextual Decision Processes},
  booktitle = {COLT},
  volume    = {99},
  pages     = {2898--2933},
  year      = {2019}
}

@inproceedings{jin2021bellmaneluder,
  author    = {Chi Jin and Qinghua Liu and Sobhan Miryoosefi},
  title     = {Bellman-Eluder Dimension: New Rich Classes of {RL} Problems, and Sample-Efficient Algorithms},
  booktitle = {NeurIPS},
  volume    = {34},
  pages     = {13406--13418},
  year      = {2021}
}

@inproceedings{du2021bilinear,
  author    = {Simon S. Du and Sham M. Kakade and Jason D. Lee and Shachar Lovett and Gaurav Mahajan and Wen Sun and Ruosong Wang},
  title     = {Bilinear Classes: A Structural Framework for Provable Generalization in {RL}},
  booktitle = {ICML},
  volume    = {139},
  pages     = {2826--2836},
  year      = {2021}
}

@inproceedings{dann2018oracle,
  author    = {Christoph Dann and Nan Jiang and Akshay Krishnamurthy and Alekh Agarwal and John Langford and Robert E. Schapire},
  title     = {On Oracle-Efficient {PAC} {RL} with Rich Observations},
  booktitle = {NeurIPS},
  year      = {2018}
}

@inproceedings{jin2020linear,
  author    = {Chi Jin and Zhuoran Yang and Zhaoran Wang and Michael I. Jordan},
  title     = {Provably Efficient Reinforcement Learning with Linear Function Approximation},
  booktitle = {COLT},
  volume    = {125},
  year      = {2020}
}

@inproceedings{yang2020kernelnn,
  author    = {Zhuoran Yang and Chi Jin and Zhaoran Wang and Mengdi Wang and Michael I. Jordan},
  title     = {Provably Efficient RL with Kernel and Neural Function Approximations},
  booktitle = {NeurIPS},
  year      = {2020}
}

@inproceedings{du2019provable,
  author    = {Simon S. Du and Akshay Krishnamurthy and Nan Jiang and Alekh Agarwal and Miroslav Dud{\'i}k and John Langford},
  title     = {Provably Efficient {RL} with Rich Observations via Latent State Decoding},
  booktitle = {ICML},
  volume    = {97},
  pages     = {1665--1674},
  year      = {2019}
}

@inproceedings{misra2020homer,
  author    = {Dipendra Misra and Mikael Henaff and Akshay Krishnamurthy and John Langford},
  title     = {Kinematic State Abstraction and Provably Efficient Rich-Observation RL},
  booktitle = {ICML},
  volume    = {119},
  pages     = {6961--6971},
  year      = {2020}
}

@inproceedings{agarwal2020flambe,
  author    = {Alekh Agarwal and Sham Kakade and Akshay Krishnamurthy and Wen Sun},
  title     = {FLAMBE: Structural Complexity and Representation Learning of Low Rank {MDP}s},
  booktitle = {NeurIPS},
  year      = {2020}
}

@inproceedings{dann2021agnostic,
  author    = {Christoph Dann and Yishay Mansour and Mehryar Mohri and Ayush Sekhari and Karthik Sridharan},
  title     = {Agnostic Reinforcement Learning with Low-Rank {MDP}s and Rich Observations},
  booktitle = {NeurIPS},
  year      = {2021}
}

@inproceedings{huang2023density,
  author    = {Audrey Huang and Jinglin Chen and Nan Jiang},
  title     = {Reinforcement Learning in Low-Rank {MDP}s with Density Features},
  booktitle = {ICML},
  volume    = {202},
  pages     = {13710--13752},
  year      = {2023}
}

@inproceedings{he2021uniformpaclinear,
  author    = {Jiafan He and Dongruo Zhou and Quanquan Gu},
  title     = {Uniform-{PAC} Bounds for RL with Linear Function Approximation},
  booktitle = {NeurIPS},
  year      = {2021}
}

@inproceedings{jin2020rewardfree,
  author    = {Chi Jin and Akshay Krishnamurthy and Max Simchowitz and Yu Tian},
  title     = {Reward-Free Exploration for Reinforcement Learning},
  booktitle = {ICML},
  volume    = {119},
  pages     = {4870--4879},
  year      = {2020}
}

@inproceedings{kaufmann2021adaptive,
  author    = {Emilie Kaufmann and Andrea Tirinzoni},
  title     = {Adaptive Reward-Free Exploration},
  booktitle = {ALT},
  volume    = {132},
  year      = {2021}
}

@inproceedings{wang2020linearRFE,
  author    = {Ruosong Wang and Simon S. Du and Lin F. Yang and Ruslan Salakhutdinov},
  title     = {On Reward-Free RL with Linear Function Approximation},
  booktitle = {NeurIPS},
  year      = {2020}
}

@inproceedings{jin2021pevi,
  author    = {Ying Jin and Zhuoran Yang and Zhaoran Wang},
  title     = {Is Pessimism Provably Efficient for Offline {RL}?},
  booktitle = {ICML},
  volume    = {139},
  year      = {2021}
}

@inproceedings{shi2022pql,
  author    = {Laixi Shi and Gen Li and Yuting Wei and Yuxin Chen and Yuejie Chi},
  title     = {Pessimistic Q-Learning for Offline RL: Towards Optimal Sample Complexity},
  booktitle = {ICML},
  volume    = {162},
  pages     = {19967--20025},
  year      = {2022}
}

@article{li2024aos,
  author  = {Gen Li and Laixi Shi and Yuxin Chen and Yuejie Chi and Yuting Wei},
  title   = {Settling the Sample Complexity of Model-Based Offline RL},
  journal = {Annals of Statistics},
  year    = {2024}
}

@article{kallus2020ope,
  author  = {Nathan Kallus and Masatoshi Uehara},
  title   = {Statistically Efficient Off-Policy Evaluation for RL},
  journal = {JMLR},
  volume  = {21},
  number  = {167},
  pages   = {1--63},
  year    = {2020}
}

@inproceedings{jiang2020minimaxinterval,
  author    = {Nan Jiang and Jiawei Huang},
  title     = {Minimax Value Interval for Off-Policy Evaluation and Policy Optimization},
  booktitle = {NeurIPS},
  pages     = {2747--2758},
  year      = {2020}
}

@inproceedings{fard2012pacbayes,
  author    = {Mahdi Milani Fard and Joelle Pineau and Csaba Szepesv{\'a}ri},
  title     = {{PAC}-Bayesian Policy Evaluation},
  booktitle = {AISTATS},
  volume    = {22},
  year      = {2012}
}

@inproceedings{rivasplata2020pacbayes,
  author    = {Omar Rivasplata and Ilja Kuzborskij and Csaba Szepesv{\'a}ri and John Shawe-Taylor},
  title     = {PAC-Bayes Analysis Beyond the Usual Bounds},
  booktitle = {NeurIPS},
  year      = {2020}
}

@article{flynn2023pacbayesbandit,
  author  = {Hamish Flynn and David Reeb and Melih Kandemir and Jan Peters},
  title   = {PAC-Bayes Bounds for Bandit Problems: A Survey and Experimental Comparison},
  journal = {IEEE TPAMI},
  year    = {2023}
}

@article{tasdighi2024pbac,
  author  = {Bahareh Tasdighi and Manuel Haussmann and Nicklas Werge and Yi-Shan Wu and Melih Kandemir},
  title   = {Deep Exploration with {PAC}-Bayes},
  journal = {arXiv preprint},
  year    = {2024}
}

@inproceedings{shi2022pomdpOPE,
  author    = {Jikai Shi and Dongruo Zhou and Quanquan Gu},
  title     = {Off-Policy Evaluation in Confounded {POMDP}s},
  booktitle = {ICML},
  volume    = {162},
  year      = {2022}
}

@inproceedings{wagenmaker2022idpac,
  author    = {Andrew J. Wagenmaker and Max Simchowitz and Kevin Jamieson},
  title     = {Beyond No-Regret: Instance-Dependent {PAC} Reinforcement Learning},
  booktitle = {COLT},
  volume    = {178},
  year      = {2022}
}

@inproceedings{tirinzoni2023optimistic,
  author    = {Andrea Tirinzoni and Mohammed Al{-}Marjani and Emilie Kaufmann},
  title     = {Optimistic {PAC} RL: the Instance-Dependent View},
  booktitle = {ALT},
  volume    = {201},
  year      = {2023}
}

@article{strehl2009pacjmlr,
  author  = {Alexander L. Strehl and Lihong Li and Michael L. Littman},
  title   = {{PAC} Analysis of a Class of RL Algorithms},
  journal = {JMLR},
  volume  = {10},
  pages   = {2413--2444},
  year    = {2009}
}

@book{szepesvari2010algorithms,
  author    = {Csaba Szepesv{\'a}ri},
  title     = {Algorithms for Reinforcement Learning},
  publisher = {Morgan \& Claypool},
  year      = {2010}
}

@article{kaelbling1996survey,
  author  = {Leslie Pack Kaelbling and Michael L. Littman and Andrew W. Moore},
  title   = {Reinforcement Learning: A Survey},
  journal = {JAIR},
  volume  = {4},
  pages   = {237--285},
  year    = {1996}
}

@book{suttonbarto2018book,
  author    = {Richard S. Sutton and Andrew G. Barto},
  title     = {Reinforcement Learning: An Introduction (2nd ed.)},
  publisher = {MIT Press},
  year      = {2018}
}

@inproceedings{azar2017ucbvi,
  author    = {Azar, Mohammad Gheshlaghi and Osband, Ian and Munos, R{\'e}mi},
  title     = {Minimax Regret Bounds for Reinforcement Learning},
  booktitle = {ICML},
  volume    = {70},
  year      = {2017}
}

@article{jaksch2010ucrl2,
  author  = {Jaksch, Thomas and Ortner, Ronald and Auer, Peter},
  title   = {Near-Optimal Regret Bounds for Reinforcement Learning},
  journal = {JMLR},
  volume  = {11},
  pages   = {1563--1600},
  year    = {2010}
}

@inproceedings{thomas2015hcope,
  author    = {Thomas, Philip S. and Theocharous, Georgios and Ghavamzadeh, Mohammad},
  title     = {High-Confidence Off-Policy Evaluation},
  booktitle = {AAAI},
  year      = {2015}
}

@inproceedings{thomas2016doublyrobust,
  author    = {Thomas, Philip S. and Theocharous, Georgios and Ghavamzadeh, Mohammad},
  title     = {Data-Efficient Off-Policy Policy Evaluation for RL},
  booktitle = {ICML},
  volume    = {48},
  year      = {2016}
}

@inproceedings{kumar2020cql,
  author    = {Kumar, Aviral and Zhou, Aurick and Tucker, George and Levine, Sergey},
  title     = {Conservative Q-Learning for Offline RL},
  booktitle = {NeurIPS},
  year      = {2020}
}

@article{valiant1984pac,
  author  = {Leslie G. Valiant},
  title   = {A Theory of the Learnable},
  journal = {Communications of the ACM},
  volume  = {27},
  number  = {11},
  pages   = {1134--1142},
  year    = {1984},
  doi     = {10.1145/1968.1972}
}

@book{bellman1957dynamic,
  author    = {Richard Bellman},
  title     = {Dynamic Programming},
  publisher = {Princeton University Press},
  year      = {1957}
}

@book{howard1960dynamic,
  author    = {Ronald A. Howard},
  title     = {Dynamic Programming and Markov Processes},
  publisher = {MIT Press},
  year      = {1960}
}

@book{puterman1994mdp,
  author    = {Martin L. Puterman},
  title     = {Markov Decision Processes: Discrete Stochastic Dynamic Programming},
  publisher = {John Wiley \& Sons},
  year      = {1994}
}

@article{papadimitriou1987complexity,
  author  = {Christos H. Papadimitriou and John N. Tsitsiklis},
  title   = {The Complexity of {M}arkov Decision Processes},
  journal = {Mathematics of Operations Research},
  volume  = {12},
  number  = {3},
  pages   = {441--450},
  year    = {1987}
}

@inproceedings{jacot2018ntk,
  author    = {Arthur Jacot and Franck Gabriel and Cl{\'e}ment Hongler},
  title     = {Neural Tangent Kernel: Convergence and Generalization in Neural Networks},
  booktitle = {NeurIPS},
  year      = {2018}
}

@article{kallus2021confounding,
  author  = {Nathan Kallus and Xiaojie Mao and Masatoshi Uehara},
  title   = {Causal Inference Under Unmeasured Confounding With Negative Controls: A Minimax Learning Approach},
  journal = {arXiv preprint arXiv:2103.14029},
  year    = {2021}
}

\appendix
\section{Notation and Abbreviations}\label{app:notation}

This appendix consolidates all notation in consolidated tables for quick reference.

\vspace{1em}
\noindent\textbf{Core MDP Notation}

\vspace{0.5em}
\begin{center}
\small
\begin{tabular}{lll}
\toprule
\textbf{Symbol} & \textbf{Meaning} & \textbf{First Defined} \\
\midrule
$\cS, \cA, \cX$ & State, action, observation/context spaces & Preliminaries \\
$S = |\cS|$, $A = |\cA|$ & Cardinalities (finite case) & Preliminaries \\
$H$ & Horizon (episode length in timesteps) & Preliminaries \\
$\gamma$ & Discount factor; $H_{\text{eff}} \approx (1-\gamma)^{-1}$ & Preliminaries \\
$\rho$ & Start-state distribution; $s_0 \sim \rho$ & Preliminaries \\
$P_h, r_h$ & Stage-$h$ transition kernel and reward & Preliminaries \\
$\pi$ & Policy (possibly nonstationary) & Preliminaries \\
$V_h^\pi, Q_h^\pi$ & Value and action-value functions & Preliminaries \\
$\pi^\star, V^\star, Q^\star$ & Optimal policy and values & Preliminaries \\
\bottomrule
\end{tabular}
\end{center}

\vspace{1.5em}
\noindent\textbf{Sample Complexity and Confidence}

\vspace{0.5em}
\begin{center}
\small
\begin{tabular}{lll}
\toprule
\textbf{Symbol} & \textbf{Meaning} & \textbf{First Defined} \\
\midrule
$\varepsilon$ & Target suboptimality (accuracy) & Definition 1 (Preliminaries) \\
$\delta$ & Failure probability (confidence $1-\delta$) & Definition 1 (Preliminaries) \\
$N(\varepsilon, \delta)$ & Episode budget for $(\varepsilon,\delta)$-PAC & Definition 1 (Preliminaries) \\
$K$ & Total episodes (regret horizon) & Preliminaries \\
$\tO(\cdot), \tTheta(\cdot), \tOmega(\cdot)$ & Suppress polylog factors & Taxonomy section \\
poly$(x)$ & Polynomial in $x$ (unspecified degree) & Throughout \\
\bottomrule
\end{tabular}
\end{center}

\vspace{1.5em}
\noindent\textbf{Structural Complexity Parameters}

\vspace{0.5em}
\begin{center}
\small
\begin{tabular}{lll}
\toprule
\textbf{Parameter} & \textbf{Meaning} & \textbf{Section} \\
\midrule
$d$ & Feature dimension (linear MDPs) & Function Approximation \\
$r$ & Low-rank factorization rank & Rich Observations \\
$m$ & Latent state size (Block MDPs) & Rich Observations \\
$B$ & Bellman rank & Structural Complexity \\
$W$ & Witness rank & Structural Complexity \\
$d_{\mathrm{BE}}$ & Bellman-Eluder dimension & Structural Complexity \\
$d_{\mathrm{eff}}(\lambda)$ & Effective dimension (RKHS/NTK) & Function Approximation \\
$\Phi_{\text{tab}}(S,A,H)$ & Tabular factor $\equiv SAH^3$ & Preliminaries \\
\bottomrule
\end{tabular}
\end{center}

\vspace{1.5em}
\noindent\textbf{Coverage and Occupancy}

\vspace{0.5em}
\begin{center}
\small
\begin{tabular}{lll}
\toprule
\textbf{Symbol} & \textbf{Meaning} & \textbf{First Defined} \\
\midrule
$d_h^\pi(s,a)$ & State-action occupancy of $\pi$ at stage $h$ & Preliminaries \\
$\mu_h(s,a)$ & Behavior/data occupancy at stage $h$ & Preliminaries \\
$C_\star$ & Concentrability coefficient & Preliminaries \\
\bottomrule
\end{tabular}
\end{center}

\vspace{1.5em}
\noindent\textbf{Function Classes and Operators}

\vspace{0.5em}
\begin{center}
\small
\begin{tabular}{lll}
\toprule
\textbf{Symbol} & \textbf{Meaning} & \textbf{First Defined} \\
\midrule
$\cF$ & Value or Q-function class & Preliminaries \\
$\cM$ & Model class (dynamics hypotheses) & Structural Complexity \\
$\cG$ & Discriminator class (witness tests) & Structural Complexity \\
$\mathcal{T}_h$ & Bellman optimality operator & Preliminaries \\
$\mathcal{N}(\epsilon, \cF, \|\cdot\|)$ & Covering number of $\cF$ & Preliminaries \\
\bottomrule
\end{tabular}
\end{center}

\vspace{1.5em}
\noindent\textbf{Access Modes}

\vspace{0.5em}
\begin{center}
\small
\begin{tabular}{ll}
\toprule
\textbf{Mode} & \textbf{Description} \\
\midrule
Online & Sequential interaction; agent chooses actions, observes transitions \\
Generative model & i.i.d.\ sampling from $P_h(\cdot \mid s, a)$ on query $(s,a,h)$ \\
Offline & Fixed dataset $\cD$; no further interaction \\
\bottomrule
\end{tabular}
\end{center}

\vspace{1.5em}
\noindent\textbf{Abbreviations}

\vspace{0.5em}
\begin{center}
\small
\begin{tabular}{ll}
\toprule
\textbf{Abbrev.} & \textbf{Meaning} \\
\midrule
RL & Reinforcement Learning \\
MDP & Markov Decision Process \\
POMDP & Partially Observable MDP \\
CDP & Contextual Decision Process \\
PAC & Probably Approximately Correct \\
RFE & Reward-Free Exploration \\
OPE & Off-Policy Evaluation \\
BPI & Best-Policy Identification \\
BE & Bellman-Eluder (dimension) \\
NTK & Neural Tangent Kernel \\
RKHS & Reproducing Kernel Hilbert Space \\
LSVI & Least-Squares Value Iteration \\
UCB & Upper Confidence Bound \\
PEVI & Pessimistic Episodic Value Iteration \\
PQL & Pessimistic Q-Learning \\
CSO & Coverage-Structure-Objective framework \\
\bottomrule
\end{tabular}
\end{center}

\end{document}